%% file: main.tex
\RequirePackage{fix-cm}
\documentclass[twocolumn]{svjour3}        
\smartqed 
\usepackage{graphicx}
\usepackage{xcolor}
\PassOptionsToPackage{hyphens}{url}\usepackage{hyperref}
\usepackage{subfigure}
\usepackage[numbers]{natbib}
\usepackage{algorithm}
\usepackage{algpseudocode}
\usepackage{multirow}
\usepackage[super]{nth}
\graphicspath{{Fig/}}

\journalname{Artificial Life and Robotics}

\begin{document}

\title{Centralization vs. decentralization in multi-robot sweep coverage with ground robots and UAVs}

\titlerunning{Centralization vs. decentralization in multi-robot sweep coverage}        

\author{Aryo~Jamshidpey \and
Mostafa~Wahby \and
Michael~Allwright \and
Weixu~Zhu \and
Marco~Dorigo \and
Mary~Katherine~Heinrich
}

\authorrunning{A. Jamshidpey et al.} 

\institute{A. Jamshidpey, M. Wahby, M. Allwright, W. Zhu, M. Dorigo, and M.K. Heinrich, \at
              IRIDIA, Universit\'{e} Libre de Bruxelles, Brussels, Belgium \\
              \email{mdorigo@ulb.ac.be, mary.katherine.heinrich@ulb.be}   
\and A. Jamshidpey \at University of Ottawa, Ottawa, Canada
\and M. Wahby \at Reverse Labs, Amsterdam, The Netherlands
}


\date{Received: date / Accepted: date}

\maketitle

\begin{abstract}
In swarm robotics, decentralized control is often proposed as a more scalable and fault-tolerant alternative to centralized control. However, centralized behaviors are often faster and more efficient than their decentralized counterparts. In any given application, the goals and constraints of the task being solved should guide the choice to use centralized control, decentralized control, or a combination of the two. Currently, the exact trade-offs that exist between centralization and decentralization are not well defined. In this paper, we compare the performance of centralization and decentralization in the example task of sweep coverage, across five different types of multi-robot control structures: random walk, decentralized with beacons, hybrid formation control using self-organizing hierarchy, centralized formation control, and predetermined. In all five approaches, the coverage task is completed by a group of ground robots. In each approach, except for the random walk, the ground robots are assisted by UAVs, acting as supervisors or beacons. We compare the approaches in terms of three performance metrics for which centralized approaches are expected to have an advantage---coverage completeness, coverage uniformity, and sweep completion time---and two metrics for which decentralized approaches are expected to have an advantage---scalability (4, 8, or 16 ground robots) and fault tolerance (0\%, 25\%, 50\%, or 75\% ground robot failure).
As expected, the results showed that the more centralized approaches greatly outperformed the decentralized ones in terms of coverage completeness, coverage uniformity, and sweep completion time. The decentralized approaches were less affected by robot failures and had better performance gains when the number of robots increased, but unexpectedly, these advantages only made their performance comparable to that of the more centralized approaches, not better than.
Finally, we discuss future work on investigating additional conditions (e.g., bottlenecks, supervisor failures, and more complex environments), and on combining the advantages of both centralization and decentralization into one system.

\end{abstract}

\section{Introduction}

When developing multi-robot systems, a fundamental design choice is whether to use centralization, decentralization, or some combination of the two. 
Ideally, one would be able to guarantee that the selected control approach is the top performer for the targeted objectives. 
However, current methods to assess performance do not span different control approaches, making direct comparisons difficult. For example, several performance metrics have been proposed for robot swarms~\citep{hamann2013towards,valentini2015efficient} and self-organizing systems \citep{kaddoum2010criteria,eberhardinger2017approach,eberhardinger2018measuring}, but these methods are particular to decentralized systems and would not apply to centralized ones.

The relative advantages and disadvantages of centralization and decentralization in multi-robot systems have been broadly referred to in the literature.
Fully centralized multi-robot systems are often high performing and efficient, but can suffer from a single point of failure or poor scalability. Fully decentralized alternatives often scale well and have some degree of inherent fault tolerance through redundancy, but can suffer from lower speed, accuracy, or efficiency compared to their fully centralized counterparts.
Despite these oft-cited general guidelines, the precise trade-offs involved when considering specific tasks are not currently known.
Some research has studied the relationship between controller structure and controller performance---e.g., in network topology~\citep{nedic2018network} or controller architectures~\citep{jovanovic2016controller}---but not the full set of issues that can arise during robot deployment, such as system fault tolerance or physical interference between robots. 

In this paper, we compare aspects of performance across different types of multi-robot control structures, using sweep coverage as an example task. We test a series of approaches spanning from more centralized to more decentralized and compare their results when tested in the same environment under the same task conditions. We define the degree of (de)centralization of a multi-robot system according to the structure of communication and style of coordination among the robots, as well as the degree to which the robots use global information about the task, the environment, or each other.

\section{Sweep coverage: related work}

Multi-robot coverage targets the systematic, uniform observation of an environment. It is relevant to adaptive sensor networks, where more efficient coverage in the exploration phase could improve the positions chosen by mobile sensors~\citep{luo2018adaptive,siligardi2019robust,santos2019decentralized}, and to robot swarms and multi-robot systems in general, where more efficient observation of the environment could improve performance during tasks such as environment monitoring, collective perception~\citep{schmickl2006collective}, search and rescue~\citep{baxter2007multi}, or foraging~\citep{lima2017cellular}.

In {\it static area coverage}, robots or sensors occupy predominantly stationary positions distributed throughout the environment~\citep{wang2011coverage}. Collectively, they monitor the entire environment without moving. In {\it sweep coverage}, by contrast, robots are mobile~\citep{galceran2013survey}. Instead of occupying stationary positions, robots move systematically through the environment and periodically visit all portions of it. In this paper, we study sweep coverage.

\subsection{Robots sweeping in parallel}

When using a single robot, the goal of sweep coverage is to quickly and efficiently collect information that comprehensively represents an environment~\citep{galceran2013survey,julia2012comparison}, for instance by sweeping the environment using boustrophedon (i.e., ``back-and-forth") motion~\citep{almadhoun2019survey,avellar2015multi}.
To speed up the process, multiple robots can sweep simultaneously.

\subsubsection{Centralized approaches}

When organized in a centralized way, the environment can be decomposed into zones which robots are assigned to sweep individually \cite[e.g.,][]{rekleitis2008efficient,scherer2015autonomous}.
If the environment is decomposed {\it a priori} and path planning is completed offline, various forms of optimization can be used~\citep{thabit2018multi,nazarahari2019multi,yu2016optimal}, often increasing efficiency but decreasing adaptability.
If decomposition and planning are instead completed online, robots can build a shared reference map of the environment either before coverage begins~\citep{mirzaei2011cooperative} or during coverage using broadcasting~\citep{miki2018multi,ge2005complete}. 
Shared reference maps are also used without strict decomposition into zones, with maps updated by single-broadcast~\citep{marjovi2009multi} or somewhat less efficiently by multi-broadcast~\citep{albani2017field}.
Approaches that update reference maps in these ways are often hybrid to some degree, rather than strictly centralized, as they include some aspects of decentralization. 

\begin{figure*}[h!]
  \centering
  \begin{minipage}[]{0.32\textwidth}
        \subfigure[\label{fig:obstacles_setup:100}
         100 obstacles]
          {\includegraphics[width=1.0\textwidth]{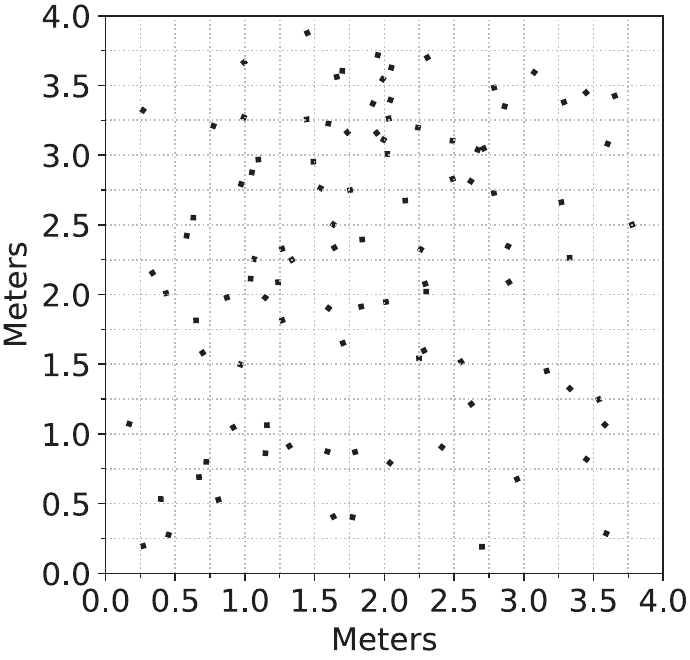}}
  \end{minipage}
  \begin{minipage}[]{0.32\textwidth}
        \subfigure[\label{fig:obstacles_setup:200}
         200 obstacles]
          {\includegraphics[width=1.0\textwidth]{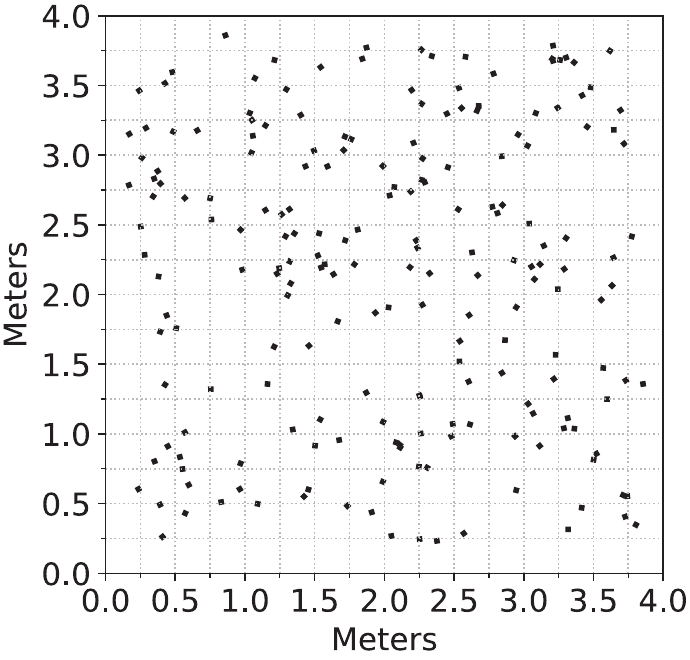}}
  \end{minipage}
  \begin{minipage}[]{0.32\textwidth}
        \subfigure[\label{fig:obstacles_setup:300}
          300 obstacles]
          {\includegraphics[width=1.0\textwidth]{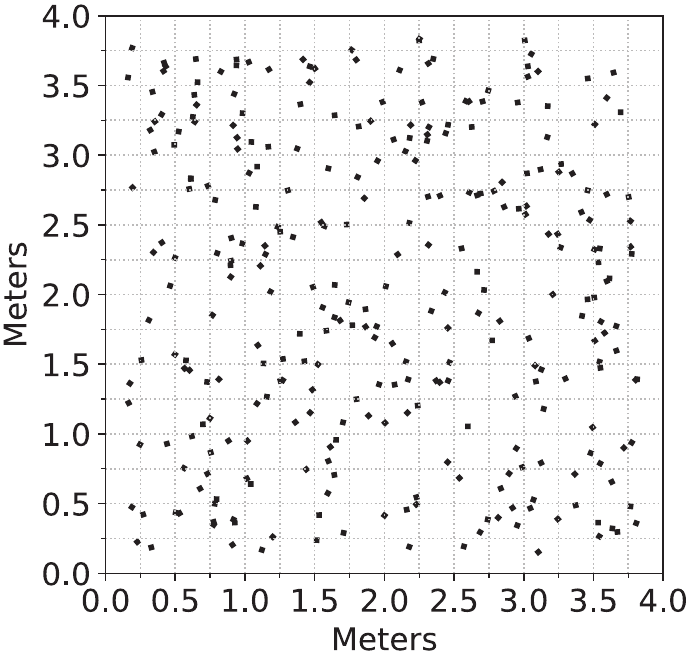}}
  \end{minipage}
  \caption{\label{fig:obstacles_setup}{\bf Illustration of obstacles in the environment.} 2D illustrations of the environment, showing the grid cells used for analysis and the number of obstacles: (a) low difficulty, 100 obstacles; (b) medium difficulty, 200 obstacles; (c) high difficulty, 300 obstacles. The obstacles (shown in black) are all the same size, but are sometimes positioned in clusters (see Sec.~\ref{sec:methods:setup} for more information). The grid cells (shown in light gray) are used only for post-hoc analysis of the experiment results; robots do not receive information about the grid cells and cannot detect them. The obstacles, arena, and grid cells are shown at their correct relative sizes. 
}
\end{figure*}

\subsubsection{Decentralized approaches}

Fully decentralized coverage can be completed using independent random walks with simple obstacle and collision avoidance~\citep{ichikawa1999characteristics,mcguire2019minimal,huang2019exploration}, but this is often highly inefficient. 
Many studies have improved upon random walk strategies, e.g., by optimizing step lengths in Brownian motion and L\'{e}vy flight~\citep{pang2021effect} or improving the scalability of L\'{e}vy walk~\citep{khaluf2018collective}.
These approaches can achieve very high coverage completeness in unknown environments~\citep[up to 97\% reported by][]{pang2021effect}, if the available time is long enough and the swarm is large enough~\citep{ichikawa1999characteristics}, even in environments cluttered with obstacles~\cite{almadhoun2019survey}.
However, inefficiency is very high, as robots often repeat coverage of areas already explored by their peers~\citep{zia2017cognitive,huang2019exploration}.

One approach to improving coordination in random walks is to leave ant-inspired artificial pheromones in the environment~\citep{koenig2001terrain}. 
Parameters of several pheromone-based approaches have been tuned for increased efficiency during sweep coverage, for example used with L\'{e}vy flight~\citep{SchRamMan2017:si} or a combination of L\'{e}vy flight and Brownian motion~\citep{deshpande2017robot}. 
A related approach leaves some robots in the environment to act as static beacons, either permanently before coverage begins~\citep{maftuleac2015local} or temporarily. In \cite{stirling2010energy}, robots temporarily park themselves as beacons, progressively forming a beacon network that spreads in a certain sub-area until that sub-area is fully explored, after which robots start re-deploying themselves in another direction, to progressively spread a beacon network in a new sub-area.
In roughly square-shaped environments with large obstacles, the approach achieved a mean coverage completeness of 99.7\% and also demonstrated scalability.

\subsection{Robots sweeping as a formation}

If robots sweep the environment as a group rather than separately, their coordination can be improved by using formation control strategies~\citep{liu2018survey} to manage motion control and relative positions. 
Formation control is often cited as a good approach for coverage applications such as search and rescue~\citep{liu2018survey,campbell2012review}, but the coverage performance of the various strategies has not been directly compared. Of the common formation shape types~\citep[cf.][]{campbell2012review}, line formations are considered suitable for sweeping and mapping tasks~\citep{liu2018survey}. 

Formation control can be fully centralized or can incorporate aspects of decentralized control, such as a potential field for local collision avoidance~\citep{liu2018survey}. 
In fully centralized approaches, robots share a common reference via broadcast, for instance a predetermined leader~\citep{wang1991navigation} or a dynamically selected navigator and virtual leader~\citep{din2018behavior}.
Formation control can also be accomplished without strictly centralized communication, for instance by using a self-organized ad-hoc communication network \citep{zhu2024self,ZhuAllHei-etal2020:ants,zhang2023self,jamshidpey2022reducing,JamZhuWah-etal2020:ants,mathews2017mergeable}.

\section{Study design}

We define the coverage task in this study as uniform and complete exploration of the environment. In other words, if an environment is partitioned into grid cells, the robots should collectively visit all cells (coverage completeness) and should visit each cell for an equal amount of time (coverage uniformity).

The experiment arena used is a $4 \times 4$~m$^2$ enclosed square, decomposed into $16 \times 16$ grid cells (grid shown in light gray in Fig.~\ref{fig:obstacles_setup}). For complete coverage, each of the 256 grid cells must be visited by the ground robots. 
The difficulty of this task is increased by adding small $4 \times 4 \times 2$~cm$^3$ obstacles to the environment (shown as small black squares in Fig.~\ref{fig:obstacles_setup}).
In each trial, the obstacle positions and orientations are randomly selected with uniform distribution.
We test arenas with the following three obstacle difficulties: 100 obstacles for low difficulty (i.e., 1\% of the environment surface occupied), 200 obstacles for medium difficulty, and 300 obstacles for high difficulty. 

This study aims to directly compare a series of multi-robot control approaches, spanning from fully decentralized to fully centralized (in terms of the communication and coordination structure as well as global versus local information). To accomplish this, we use heterogeneous multi-robot systems, differentiating between robots that perform the coverage task (ground robots) and robots that instead supervise (unmanned aerial vehicles –– UAVs). The ground robots are also responsible for their own local obstacle avoidance. In all approaches, the number of ground robots and their hardware capabilities remain constant, while the number and capabilities of the UAVs depend on the approach being used. Note that UGV (unmanned ground vehicles) supervisors could in principle be used instead of UAVs, but they would need to have additional capabilities compared to the non-supervisors, so the system would still be heterogeneous.
We test the following five approaches: 
\begin{enumerate}
    \item \textit{Random walk.} ~~The most extreme form of decentralized sweep coverage found in the literature is random walk, which operates without any communication or guidance. Despite its simplicity and inefficiency, random walk has been shown to be capable of very high coverage completeness when allowed to run for sufficient time~\citep[up to 97\% completeness reported by][]{pang2021effect}. In this study, we use random billiards (i.e., ballistic motion), which has previously shown superior performance compared to other random walks, such as Brownian motion and L\'{e}vy walk \citep{kegeleirs2019random}. Additionally, in our preliminary testing (see Table 1 in the Appendix), it achieved the highest coverage completeness among the tested random walks, comparable to that of the tested stigmergic approach. Since random billiards operate without communication or guidance, the approach does not rely on UAVs, which are therefore not used. In this paper, the random billiards approach can serve as a performance benchmark.
    \vspace{2mm}
    \item \textit{Decentralized with beacons.} ~~One of the most successful approaches in the literature to improve coordination and efficiency in decentralized coverage control is to deploy robots as static beacons~\citep[mean completeness of 99.7\% reported by][]{stirling2010energy}. In this study, the UAVs serve as beacons, communicating with the ground robots during sweep coverage to minimize redundant cell visits. 
    \vspace{2mm}
    \item \textit{Hybrid formation.} ~~To increase coordination among robots further, they can perform sweep coverage while moving together in a line formation~\citep{liu2018survey}. In this study, for a formation approach that does not use fully centralized communication, we use coordination through a self-organizing hierarchy~\citep{zhu2024self}. Ground robots operate at the lowest hierarchy level, receiving motion instructions from multiple UAV supervisors that occupy the hierarchy level above them. The UAVs have limited fields of view and can only communicate with the ground robots within their detection range. The UAVs sweep the environment reactively, by detecting and reacting to its enclosing walls. 
    \vspace{2mm}
    \item \textit{Centralized formation.} ~~The fully centralized formation approach is similar to the hybrid formation approach, but uses a single UAV supervisor that has a global view of the entire environment and can communicate directly with all ground robots.
    \vspace{2mm}
    \item \textit{Predetermined.} ~~The most extreme form of centralized sweep coverage in the literature uses \textit{a priori} knowledge to design and optimize motion control offline. The environment is decomposed offline, robots are assigned to sections to sweep independently, and the paths of those robots are then planned offline. After planning is complete, sweep coverage is executed. In this paper, the predetermined decomposition and target paths are known by a UAV supervisor, which gives instructions to the ground robots as they sweep. The ground robots remain responsible for their own obstacle avoidance. The motion routine they use is designed to ensure that ground robots will circumnavigate obstacles and return to the predetermined target path, based on \textit{a priori} knowledge of the obstacle size and type of distribution, as well as the type of sweeping path the UAV will calculate.
\end{enumerate}

\begin{figure*}[h!]
\centering
\includegraphics[width=0.82\textwidth]{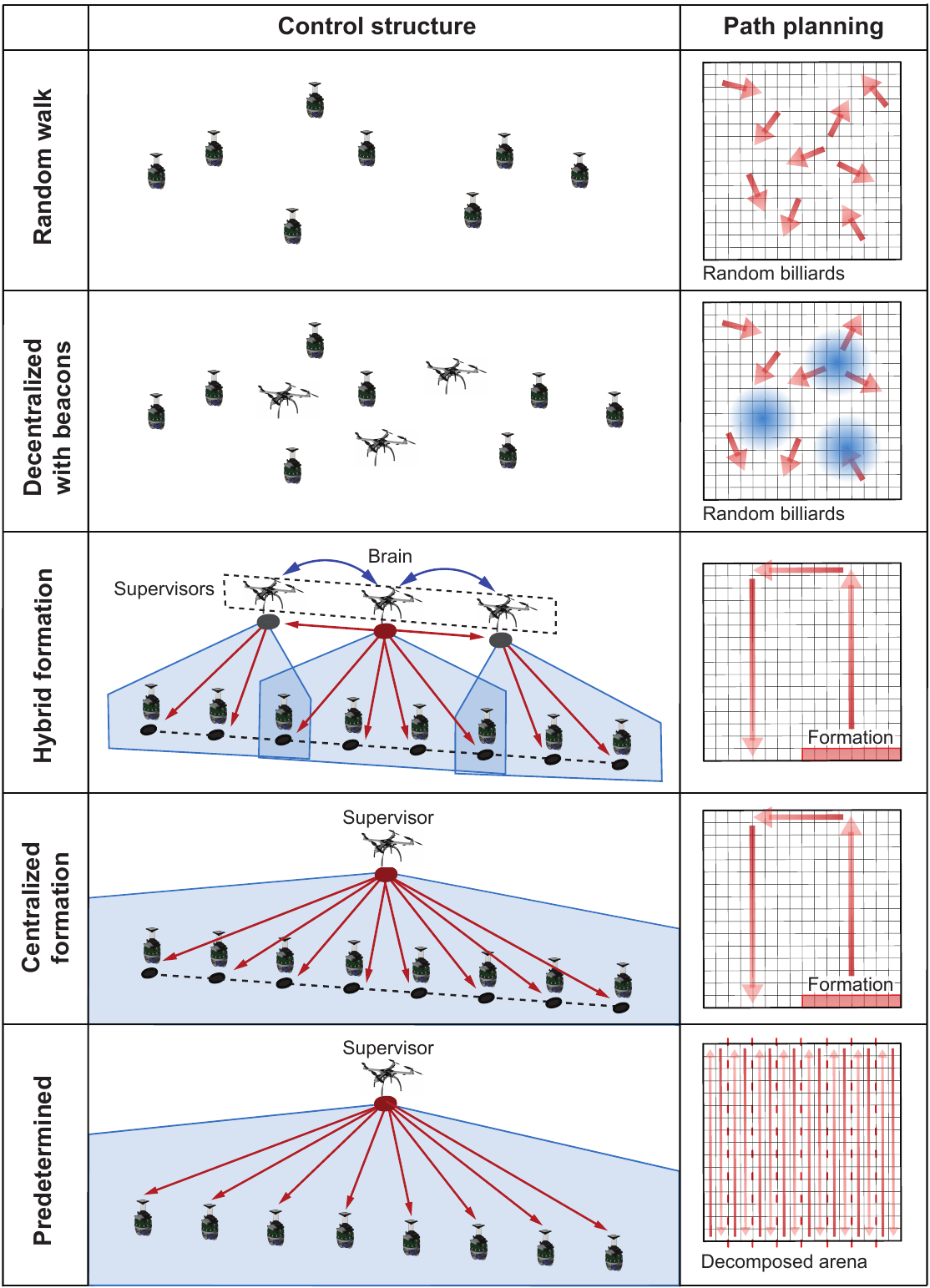}
\caption{The five control approaches. {\bf Random walk:} Ground robots explore the environment independently via randomized motion control, without any UAV supervisors. 
{\bf Decentralized with beacons:} Ground robots explore the environment independently via randomized motion control, except when UAV beacons instruct them to vacate an area.
{\bf Hybrid formation:} The UAV ``brain'' conducts a reactive boustrophedon sweep and all UAVs give motion instructions to ground robots in their limited fields of view, using a self-organized hierarchical communication network (using the SoNS). {\bf Centralized formation:} One UAV with unlimited field of view conducts a reactive boustrophedon sweep and gives motion instructions to all ground robots using a predefined centralized communication network. {\bf Predetermined:} After the environment is decomposed and paths are planned offline, one UAV gives ground robots predetermined motion instructions, using a predetermined centralized communication network. 
}
\label{fig:4_approaches}
\end{figure*}

\section{Methods}

\subsection{Random walk}

As a benchmark of fully decentralized control with no UAV supervision, we use a random walk based on random billiards~\citep{comets2009billiards}. Each ground robot independently moves at a constant velocity, except when reflecting off a boundary in a random direction or turning to avoid an object, which could be an obstacle or another robot. When avoiding an object, a robot rotates until the object is no longer in its heading direction. The pseudocode for boundary reflection and object avoidance is given in Algorithm~\ref{alg:decentral}.
The robots' initial positions and headings are randomly assigned following a uniform distribution.

\subsection{Decentralized with beacons}

In this approach, the default behavior of the ground robots is the same as in the random walk, but they can also receive guidance from UAVs that act as static beacons in the environment. 
In each trial, the positions and orientations of the UAV beacons are randomly assigned, ensuring a 0.5 m buffer from the outer boundary and a 1.5 m buffer between UAVs. The UAV beacons know the total number of ground robots in the system, but have a limited field of view for both detection and communication. While a UAV detects more than 25\% of the total ground robots within its field of view, it sends messages to all the robots it detects, instructing them to move away (until the UAV no longer detects more than 25\% of the total ground robots). The message also informs the recipient of the recipient's position and heading relative to the UAV. When a ground robot receives a message requesting departure, it turns to face away from the respective UAV, then resumes its default behavior. The pseudocode for this behavior is given in Algorithm~\ref{alg:beacon}.

\subsection{Hybrid formation based on the \textit{self-organizing nervous system} (SoNS)}

We use hybrid formation control based on the \textit{self-organizing nervous system} (SoNS)~\cite{zhu2024self}. Using the SoNS, a heterogeneous swarm can self-organize into a dynamic ad-hoc communication network with a hierarchical structure in the form of a tree (the root of the tree is called \emph{brain}, and the nodes directly connected to the brain are its children; for full details, see~\cite{zhu2024self}). Using this hierarchy, robots report sensing events and cede part of their autonomy to the brain robot (see Fig.~\ref{fig:4_approaches}). The brain robot determines its own motion trajectory and, as it moves, it acts as a motion reference for its children (i.e., the brain acts as a reference coordinate frame used to calculate the motion instructions sent to its children, which in turn act as motion references for their children). The robots in a SoNS can self-reconfigure into a new communication network and control hierarchy on the fly, for instance, in case of task change or brain failure~\citep{zhu2024self}.

\begin{algorithm}[t!]
\caption{{\bf - Ground robot behavior:} Random walk.}
\label{alg:decentral}
\begin{algorithmic}[1]
  \If{an environment boundary is detected}
  \State \parbox[t]{210pt}{turn to a random direction facing away from the boundary\strut}
  \ElsIf{an object is detected within 60$^{\circ}$ of the heading}{
          \If{the object is located on the lefthand side}
          \State turn right
       \Else~turn left \EndIf}
   \Else~move forward \EndIf
\end{algorithmic}
\end{algorithm}
\begin{algorithm}[t]
\caption{{\bf - Ground robot behavior:} Decentralized with beacons.}
\label{alg:beacon}
\begin{algorithmic}[1]
  \State {\bf update} the \textsc{rotation direction}, which is set randomly at initialization to either clockwise or counterclockwise, then alternates every $k_1$ steps
  \If{a message requesting departure is received from a beacon and the beacon is within 90$^{\circ}$ of the heading}
  \State \parbox[t]{210pt}{rotate either clockwise or counterclockwise, according to the current \textsc{rotation direction}\strut}
  \Else~follow \textbf{Algorithm~\ref{alg:decentral}} \EndIf
\end{algorithmic}
\end{algorithm}
\begin{algorithm}[t]
\caption{{\bf - Ground robot behavior:} Hybrid formation and centralized formation methods. For the accompanying UAV behaviors, see flowcharts in Figs.~\ref{fig:Flow_SoNS_UAV} and~\ref{fig:Flow_CentForm_UAV}.}
\label{alg:formations}
\begin{algorithmic}[1]
  \If{an object is detected within 60$^{\circ}$ of the heading}{
        \If{the object is located on the lefthand side}
        \State turn right
        \Else~turn left \EndIf}
  \ElsIf{an object is detected within 90$^{\circ}$ of the heading}{
       \If{the object is located on the lefthand side}
       \State move forward while turning left
       \Else~move forward while turning right \EndIf}
  \Else~follow motion instructions from UAV~~~~{\it (See Fig.~\ref{fig:flow})}  \EndIf
\end{algorithmic}
\end{algorithm}
\begin{algorithm}[t]
\caption{{\bf - Ground robot behavior:} Predetermined. (See {Alg.~5} in the Appendix for more details.) For the accompanying UAV behavior, see flowchart in Fig.~\ref{fig:Flow_FullCent_UAV}.}
\label{alg:central}
\begin{algorithmic}[1]
  \If{an object is detected directly in front of the heading or slightly to the left}
  \State \parbox[t]{210pt}{follow a pre-designed motion routine resulting in a roughly half-circle trajectory around the object(s) {\bf until} the UAV sends instructions to stop\strut}
  \ElsIf{an object is detected on the lefthand or righthand side}
  \State \parbox[t]{210pt}{turn slightly away from the object while moving forward\strut}
  \Else~follow motion instructions from UAV~~~~{\it (See Fig.~\ref{fig:flow})} \EndIf
\end{algorithmic}
\end{algorithm}

\begin{figure*}[t]
  \centering
  \begin{minipage}[t]{0.33\textwidth}
        \vspace{25mm}
        \subfigure[\label{fig:Flow_SoNS_UAV}
        {\bf UAV behavior:} Hybrid formation (brain and non-brain UAVs), with boustrophedon sweep behavior highlighted in blue. For the accompanying ground robot behavior, see Alg.~\ref{alg:formations}.]
        {\includegraphics[width=0.99\textwidth,trim=-5mm 0 -5mm 0,clip]{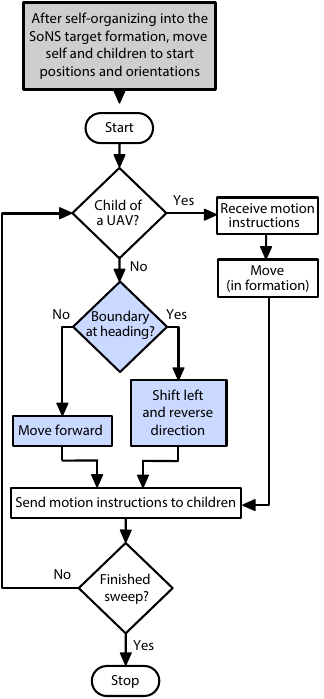}}
  \end{minipage}
  \hspace{15mm}
  \begin{minipage}[t]{0.33\textwidth}
	\centering
        \subfigure[\label{fig:Flow_CentForm_UAV}
        {\bf UAV behavior:} Centralized formation, with boustrophedon sweep behavior highlighted in blue. For the accompanying ground robot behavior, see Alg.~\ref{alg:formations}.\vspace{2mm}]
        {\includegraphics[width=0.8\textwidth,trim=-5mm 0 -5mm 0,clip]{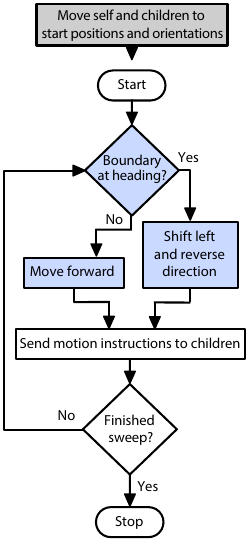}}
        \subfigure[\label{fig:Flow_FullCent_UAV}
        {\bf UAV behavior:} Predetermined. For the accompanying ground robot behavior, see Alg.~\ref{alg:central}. \vspace{2mm}]
        {\includegraphics[width=0.8\textwidth,trim=-9.5mm 0 -9.5mm 0,clip]{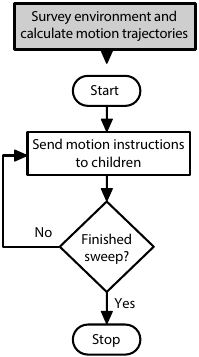}}
  \end{minipage}
  \caption{\label{fig:flow}
\textbf{UAV behaviors.} Flowcharts for the UAV behaviors in the (a) hybrid formation, (b) centralized formation, and (c) predetermined methods. In the hybrid formation and centralized formation methods, the boustrophedon sweep behavior of the UAVs is the same (see portion of the flowchart highlighted in blue): in both cases, the boustrophedon sweep is accomplished simply by shifting left and reversing direction when a boundary is met. In both cases, the dimensions of the target formation are fixed, and the UAVs are aware of those dimensions, so they shift approximately one formation-length to the left.}
\end{figure*}

In our hybrid formation setup, the target communication network topology is a caterpillar tree---i.e., a tree in which all inner nodes are on one central path, to which each leaf node is connected.
UAVs try to recruit ground robots and assign them to leaf nodes in the communication network. The ground robots complete the coverage task using instructions they receive from UAV supervisors and perform
independent obstacle avoidance. The pseudocode for the ground robots is given in Algorithm~\ref{alg:formations}; note that if ground robots temporarily lose connection with their UAV supervisors, they revert to the random walk behavior.
UAV supervisors try to recruit each other, such that two of them become inner nodes and one becomes the brain node. The UAVs act as motion references for their children and direct their ground robot children into a line formation, as shown in Fig.~\ref{fig:4_approaches}, and prevent collisions between ground robots.
The UAV that becomes the brain~\citep[see][]{zhu2024self} reactively sweeps the environment using back-and-forth boustrophedon motions~\citep{choset1998coverage}, in the same manner as in~\citep{jamshidpey2022reducing}.
When the brain detects an arena boundary in front of it and shifts to the left, its motion calculation is based on its knowledge of the formation's dimensions. Thus, the ground robots' coverage of the environment is maximized regardless of how many ground robots are in the formation.
A behavior flowchart for all UAVs in the SoNS hybrid formation is given in Fig.~\ref{fig:Flow_SoNS_UAV}.

When not hovering, the brain UAV moves at a constant speed, regardless of the speed of the other robots. The formation is maintained because the other robots adjust their motion according to the motion of their parents.
The brain's reactive boustrophedon sweep is deterministic, such that one sweep of the environment in the same conditions always takes the same amount of time. 

\subsection{Centralized formation}

For centralized formation control, we use a basic leader--follower approach \citep{wang1991navigation}, in which all robots follow one global leader that broadcasts information.
In our approach, the ground robots are connected to a UAV hub through a star-shaped (i.e., hub-and-spoke) communication network. The UAV acts as a single coordinating entity, providing all ground robots with motion instructions. 
The UAV's instructions keep the ground robots in the target line formation, as shown in Fig.~\ref{fig:4_approaches}, and prevent collisions between ground robots. The UAV follows the same reactive boustrophedon sweep used by the brain in the SoNS hybrid formation. A behavior flowchart for the UAV is given in Fig.~\ref{fig:Flow_CentForm_UAV}.
The ground robots perform independent obstacle avoidance (see pseudocode in Algorithm~\ref{alg:formations}), but otherwise follow the UAV's instructions.

\subsection{Predetermined control based on {\it a priori} knowledge}

For predetermined coverage control, we decompose the environment offline into zones for robots to sweep individually~\citep[e.g.,][]{rekleitis2008efficient,scherer2015autonomous}.
As in the centralized formation approach, ground robots receive instructions from a single UAV through a predetermined central communication network. Unlike the centralized formation approach, the UAV does not use reactive control to update the target sweep paths in real time---rather, the UAV relies on {\it a priori} knowledge to decompose the environment and calculate predetermined target sweep paths for the ground robots before they begin the task, then gives them instructions as they sweep, to keep them on the predetermined target paths.
The predetermined target paths for the ground robots to sweep their zones are based on boustrophedon motion~\citep[cf.][]{avellar2015multi}. A behavior flowchart for the UAV is given in Fig.~\ref{fig:Flow_FullCent_UAV}.

The ground robots are given a motion routine to circumnavigate obstacles in a roughly half-circle (see pseudocode in Algorithm~\ref{alg:central}), which is pre-designed to work well with a boustrophedon sweep and with the utilized obstacle size and type of distribution.
In this approach, the UAV supervisor is assumed to have complete knowledge of the environment shape, size, and boundaries; the dimensions and layout of the grid cells; and the number, positions, and orientations of the robots. It is connected to all ground robots as the hub of a star network (as in the centralized formation approach). Before the start of the experiment, the environment is decomposed into lanes that are suitable for the number of ground robots, the boustrophedon motion style, the dimensions of the environment, and the dimensions of the grid cells (see Fig.~\ref{fig:4_approaches}), and the ground robots are given ideal starting positions at the corners of their respective lanes. 

\subsection{Experiment setup}
\label{sec:methods:setup}

The experiments were conducted in the ARGoS simulator~\citep{pinciroli2012argos} using plugins~\citep{allwright2018argos,allwright2018simulating} for the ground robot and UAV models.
The simulated ground robot is based on the e-puck~\citep{mondada2009puck}. It uses differential drive with an average speed of $7$~cm/s and is equipped with a ring of 12 short-range proximity sensors.
The ground robots can detect objects (both obstacles and other ground robots) at a distance of up to 3 to 5~cm and can detect an environment boundary when on top of it.
The simulated UAV is a quadrotor equipped with a downward-facing camera (able to detect ground robots and environment boundaries) and moves at a speed of $7$~cm/s when not hovering, so that the slower ground robots can follow the UAVs. Obstacles cannot be detected by the UAVs, only by the ground robots.
The bodies of both robot types are represented by simple $2.5$~cm radius cylinders.

Fiducial markers encoding unique IDs sit atop the ground robots and can be detected by the UAVs. The ground robots are incapable of localizing themselves, but the UAVs can track the relative positions and orientations of the ground robots in order to give them motion instructions.
In the centralized formation and predetermined approaches, there is one supervisor UAV that can view the full arena and communicate with all ground robots. 
In the SoNS hybrid formation approach, a UAV can communicate with a ground robot if the ground robot is in its field of view, and two UAVs can communicate if they can see the same ground robot~\citep[for details, see][]{ZhuAllHei-etal2020:ants}.
The three UAVs fly at a constant altitude, and each UAV can view ground robots in an approximately $1.65 \times 1.85$~m$^2$ rectangular ground area~\citep[for details, see][]{ZhuAllHei-etal2020:ants}. 

At the start of each experiment, the $4 \times 4 \times 2$~cm$^3$ obstacles are distributed randomly throughout the $4 \times 4$~m$^2$ arena (shown as small black squares in Fig.~\ref{fig:obstacles_setup}), but with obstacles that are very close to each other getting clustered side-by-side (in clusters not exceeding $k_2$ in number), and while ensuring at least a $20$~cm buffer between obstacles or obstacle clusters on at least one axis of the environment coordinate system ($x$ or $y$) and at least a $18$~cm buffer between obstacles and the outer boundary. The buffers prevent the ground robots from becoming stuck.

In all approaches, the point at which the robots start the coverage task is considered to be the start of the experiment. The time needed for the robots to get into their starting positions (e.g., the time to randomly distribute the ground robots or beacons, or to connect the robots to their supervisor and establish the desired formation shape) is not included in the experiment time.

For the hybrid formation, centralized formation, and predetermined approaches, experiments are terminated at the completion of a sweep. A sweep takes the same amount of time in both the hybrid and centralized formation approaches because the UAVs in both cases use the same deterministically reactive boustrophedon sweep. The random walk and decentralized approach with beacons do not perform a sweep with a clearly defined endpoint. Instead, in these approaches, a ``sweep'' is considered complete when no robot has discovered an unvisited cell in the last $k_3$ time steps. This termination condition prevents coverage uniformity from worsening while coverage completeness is stagnating.

\begin{figure*}
  \centering
  \begin{minipage}[t]{0.42\textwidth}
        \subfigure[\label{fig:time_100_mixed}
         Low difficulty: 100 obstacles.]
          {\includegraphics[width=1.0\textwidth]{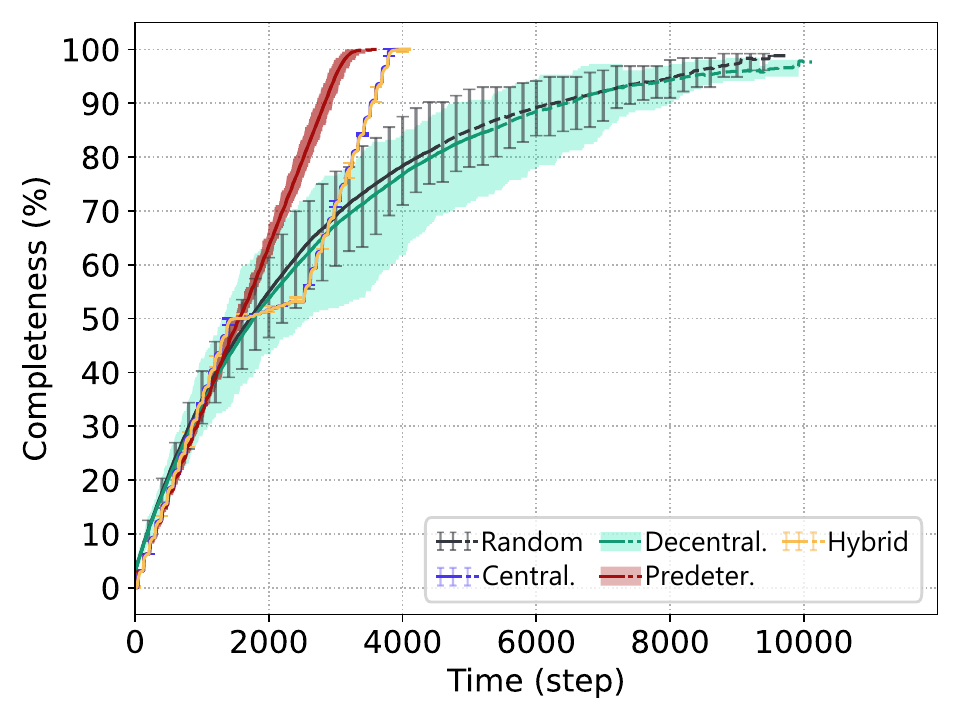}}
  \end{minipage}
  \begin{minipage}[t]{0.42\textwidth}
        \subfigure[\label{fig:time_200_mixed}
         Medium difficulty: 200 obstacles.]
          {\includegraphics[width=1.0\textwidth]{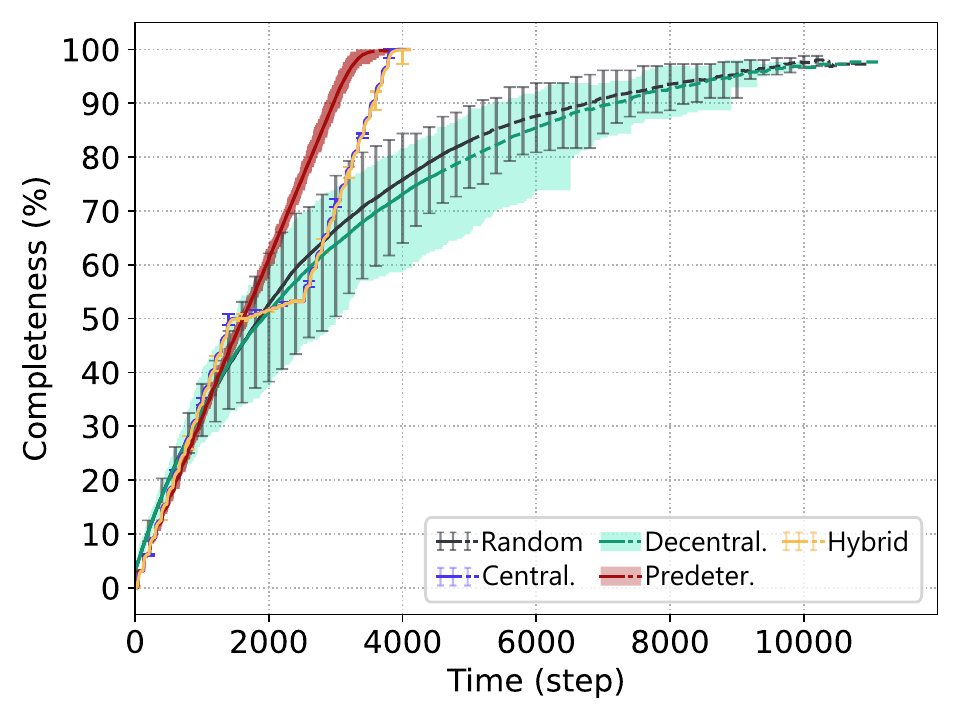}}
  \end{minipage}
    \begin{minipage}[t]{0.42\textwidth}
        \subfigure[\label{fig:time_300_mixed}
         High difficulty: 300 obstacles.]
          {\includegraphics[width=1.0\textwidth]{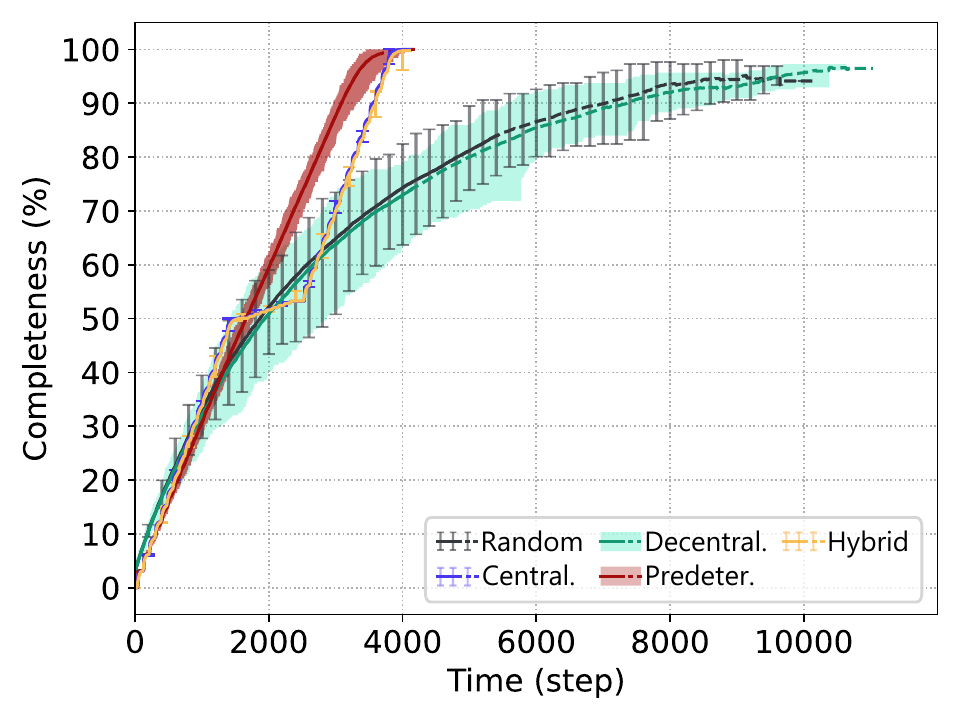}}
  \end{minipage}
  \caption{\textbf{Performance: Coverage completeness over time.} Shaded area or bars represent the minimum to maximum; lines represent the mean; dotted lines indicate that some of the trials have completed the sweep, so are no longer included in the data for that time step. Random walk (gray), decentralized with beacons (green), hybrid formation (yellow), centralized formation (purple), and predetermined (red) approaches.}
  \label{fig:perf-time}
\end{figure*}

\section{Results}
\label{sec:results}

To assess the impact of (de)centralization on coverage control, we run experiments that test the coverage performance of all five approaches: random walk, decentralized with beacons, hybrid formation, centralized formation, and predetermined.

In high-performing and efficient multi-robot coverage, robots cover a high percentage of the environment with a low rate of repeated coverage. In the performance experiments, we assess the performance of the five approaches in terms of coverage completeness (i.e., the percentage of grid cells visited) and coverage uniformity (i.e., the variability of the time robots collectively spent visiting each cell), and sweep completion time, in each of the three obstacle setups (100, 200, and 300 obstacles), with each approach using 8 ground robots (the number of UAVs varies by approach).
We also conduct fault tolerance and scalability experiments, in which we assess the effect of a percentage of failed ground robots (0\%, 25\%, 50\%, and 75\% of a total of 8 ground robots) and of the total number of ground robots (4, 8, and 16 ground robots) on the performance metrics (coverage completeness and uniformity, and sweep completion time), in the setup with 100 obstacles. 
We complete 50 trials for each type of control structure in each setup, for a total of 2\,000 trials.

We report the experimental data in Figs.~\ref{fig:perf-time}—\ref{fig:scalability}, in tabular form in the Appendix, and in an open-access data repository\footnote{\url{https://doi.org/10.5281/zenodo.14846074}}. 

\begin{figure*}
  \centering
  \begin{minipage}[t]{0.43\textwidth}
        \subfigure[\label{fig:Completeness}
         Coverage completeness.]
          {\includegraphics[width=1.0\textwidth]{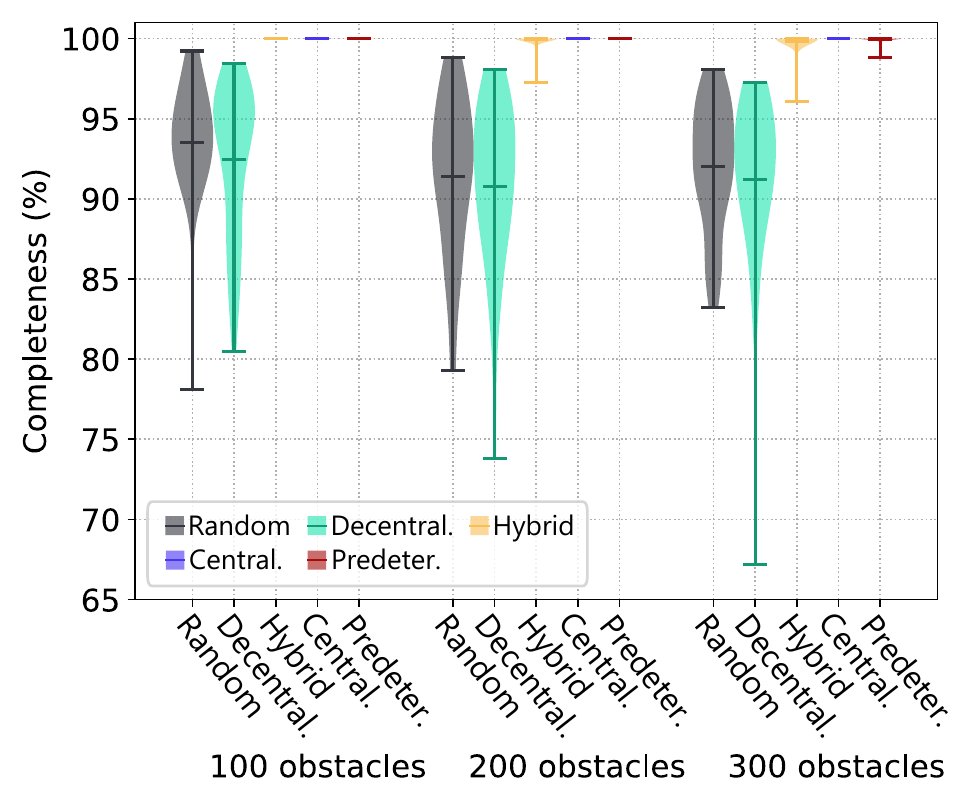}}
  \end{minipage}
  \begin{minipage}[t]{0.43\textwidth}
        \subfigure[\label{fig:Uniformity}
         Coverage uniformity.]
          {\includegraphics[width=1.0\textwidth]{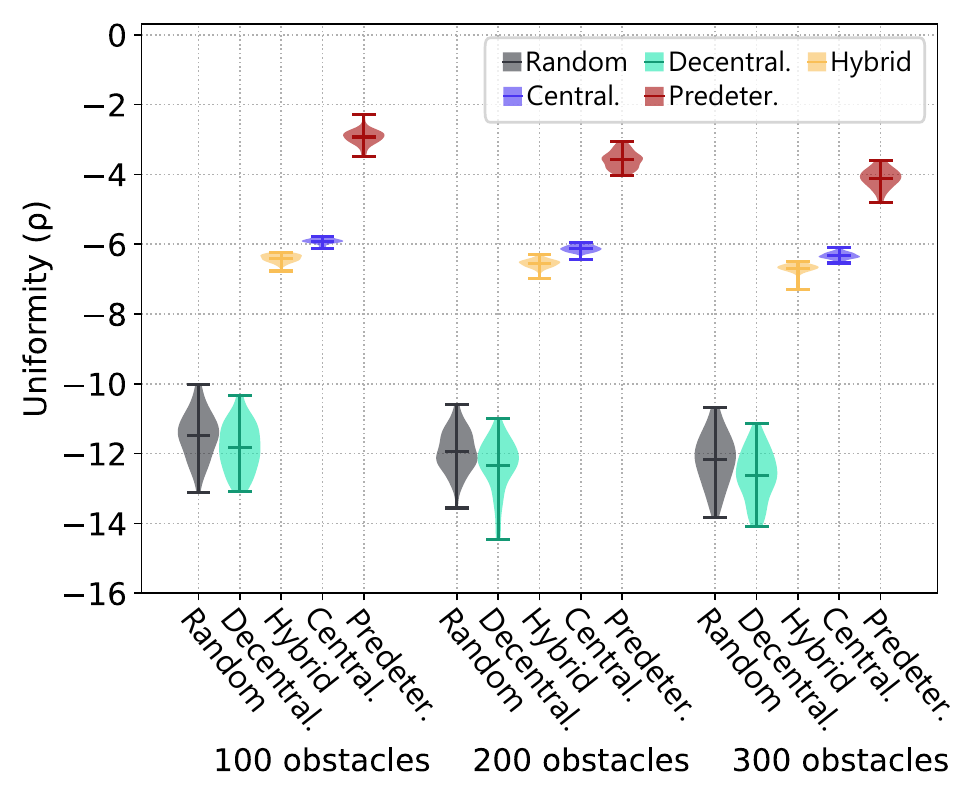}}
  \end{minipage}
  \caption{\textbf{Performance: Overall coverage completeness and uniformity.} Random walk (gray), decentralized with beacons (green), hybrid formation (yellow), centralized formation (purple), and predetermined (red) approaches.}
  \label{fig:perf-overall}
\end{figure*}

\subsection{Performance} 
\subsubsection{Coverage completeness}
\label{sec:results:completeness}

The coverage completeness (i.e., the percentage of grid cells visited) results are presented in Figs.~\ref{fig:perf-time} and~\ref{fig:perf-overall}, as well as Tables 2 and 3 in the Appendix.

At the end of a sweep, the predetermined, centralized formation, and hybrid formation approaches all achieved perfect or nearly perfect coverage completeness (99.9\% to 100\% completeness on average), with slightly more outlier variation occurring in the hybrid formation approach (see Fig.~\ref{fig:perf-overall}a). The predetermined approach, however, was faster than the centralized and hybrid formation approaches, most noticeably in the condition with 100 obstacles. As the number of obstacles increased, the time advantage of the predetermined approach decreased (taking approximately 17\% less time on average with 100 obstacles, versus approximately 6\% less time on average with 300 obstacles).
The decentralized approaches are very similar in terms of average coverage completeness and completeness variability, with the decentralized approach with beacons showing slightly greater deviations.
Both decentralized approaches achieve noticeably lower completeness than the predetermined and formation approaches and noticeably greater completeness variability, but they still perform quite well (91.5\% to 92.3\% completeness on average). However, they take approximately twice as much time to complete a sweep.
For all approaches, the number of obstacles has no clear effect on the average coverage completeness achieved.

\begin{figure*}[h!]
\centering
    \subfigure[\label{fig:time_0failuer_mixed}
         No failure.]
          {\includegraphics[width=0.42\textwidth]{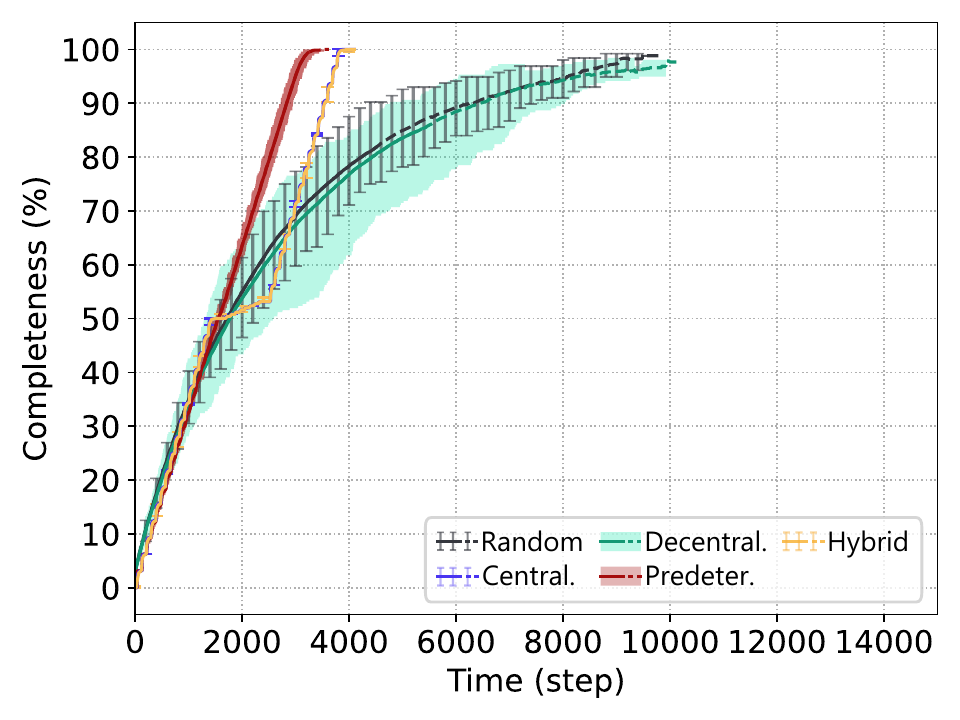}}
          \hspace{4mm}
    \subfigure[\label{fig:time_2failuers_mixed}
         2 failures (25\% failure).]
          {\includegraphics[width=0.42\textwidth]{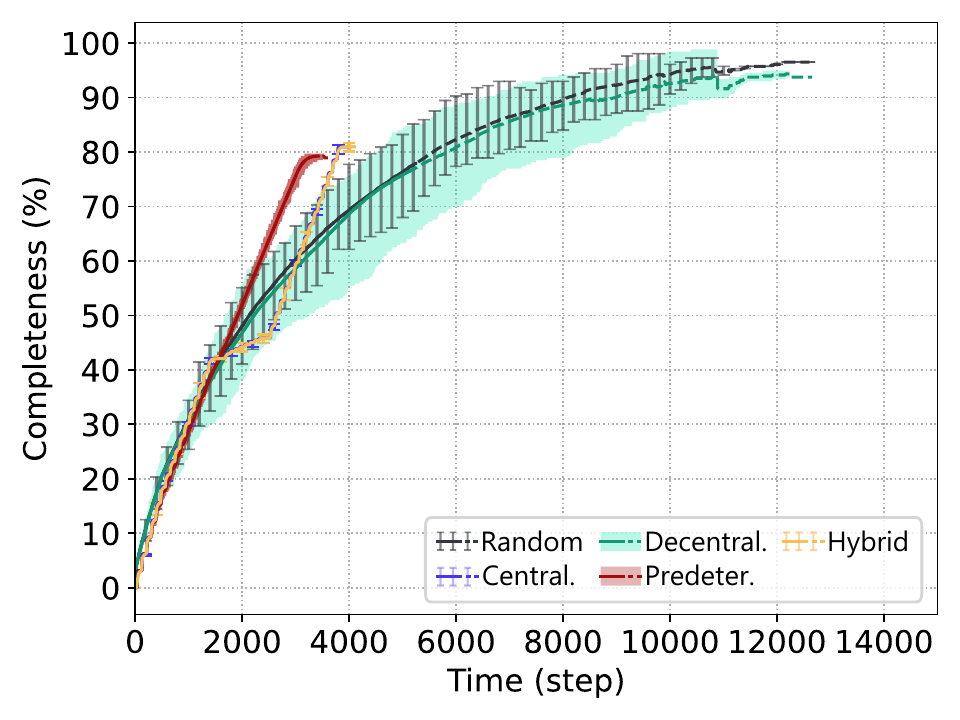}}
    \subfigure[\label{fig:time_4failuers_mixed}
         4 failures (50\% failure).]
          {\includegraphics[width=0.42\textwidth]{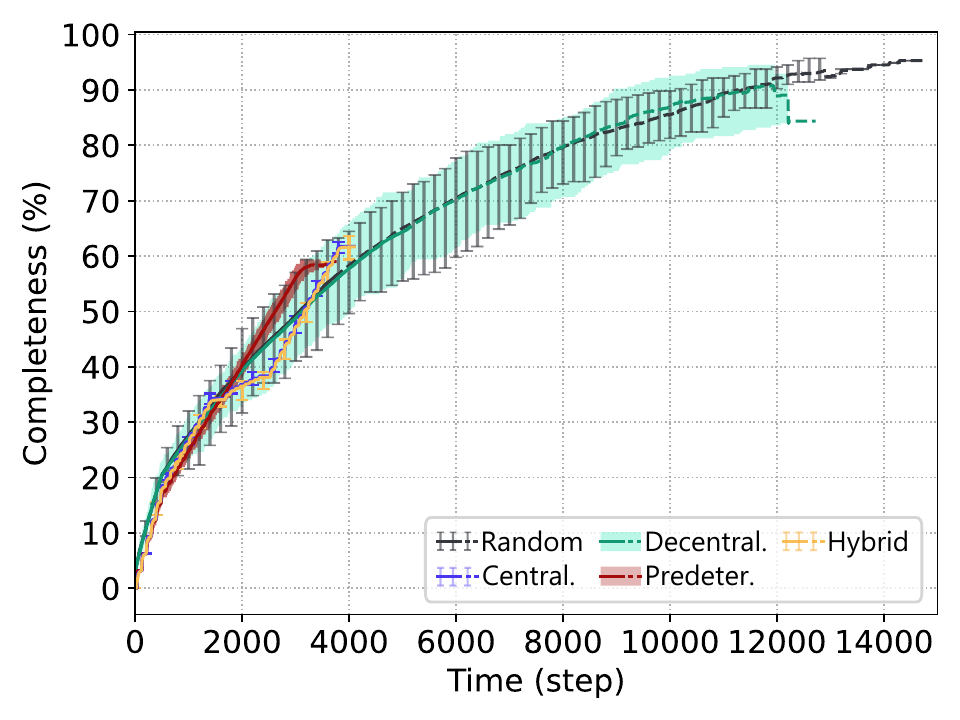}}
          \hspace{4mm}
    \subfigure[\label{fig:time_6failuers_mixed}
         6 failures (75\% failure).]
          {\includegraphics[width=0.42\textwidth]{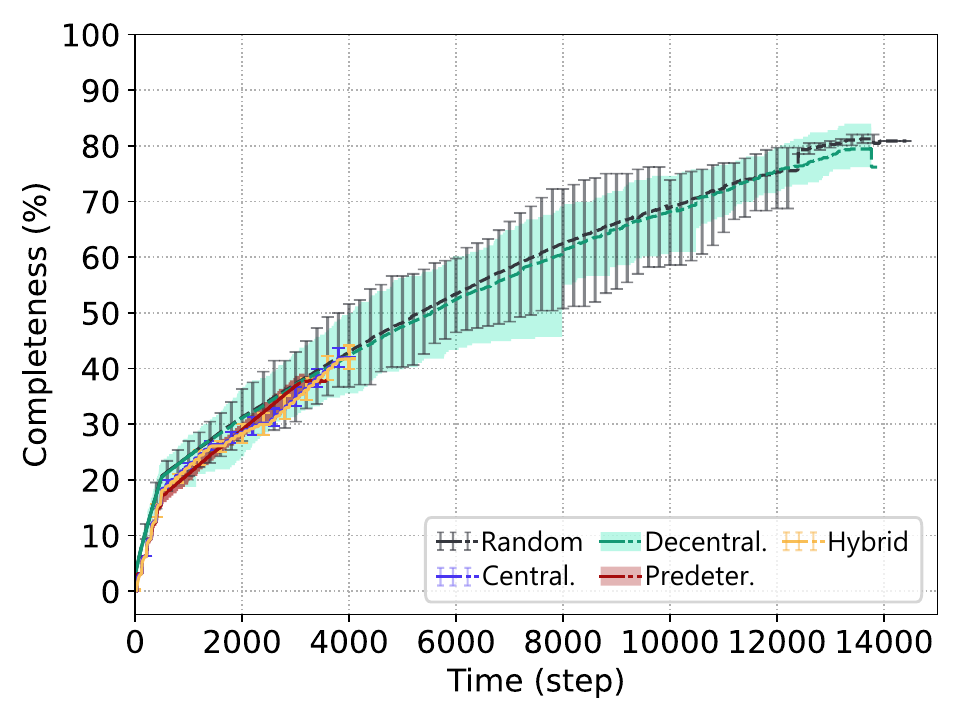}}
  \caption{\textbf{Fault tolerance: Coverage completeness over time under varying loss conditions.} Shaded area or bars represent the minimum to maximum; lines represent the mean; dotted lines indicate that some of the trials have completed the sweep, so are no longer included in the data for that time step. Random walk (gray), decentralized with beacons (green), hybrid formation (yellow), centralized formation (purple), and predetermined (red) approaches.}
  \label{fig:fault-time}
\vspace{2mm}
    \subfigure[\label{fig:FaultTolerance_Completeness}
         Fault tolerance: Coverage completeness.]
          {\includegraphics[width=0.43\textwidth]{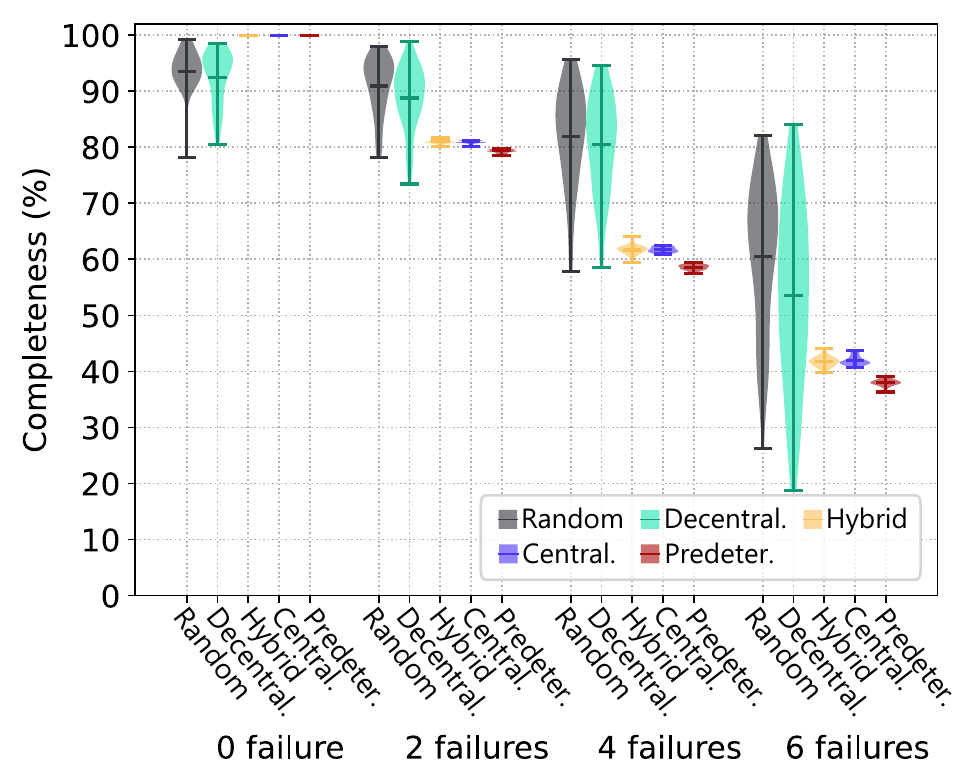}}
    \subfigure[\label{fig:FaultTolerance_Uniformity}
         Fault tolerance: Coverage uniformity.]
          {\includegraphics[width=0.43\textwidth]{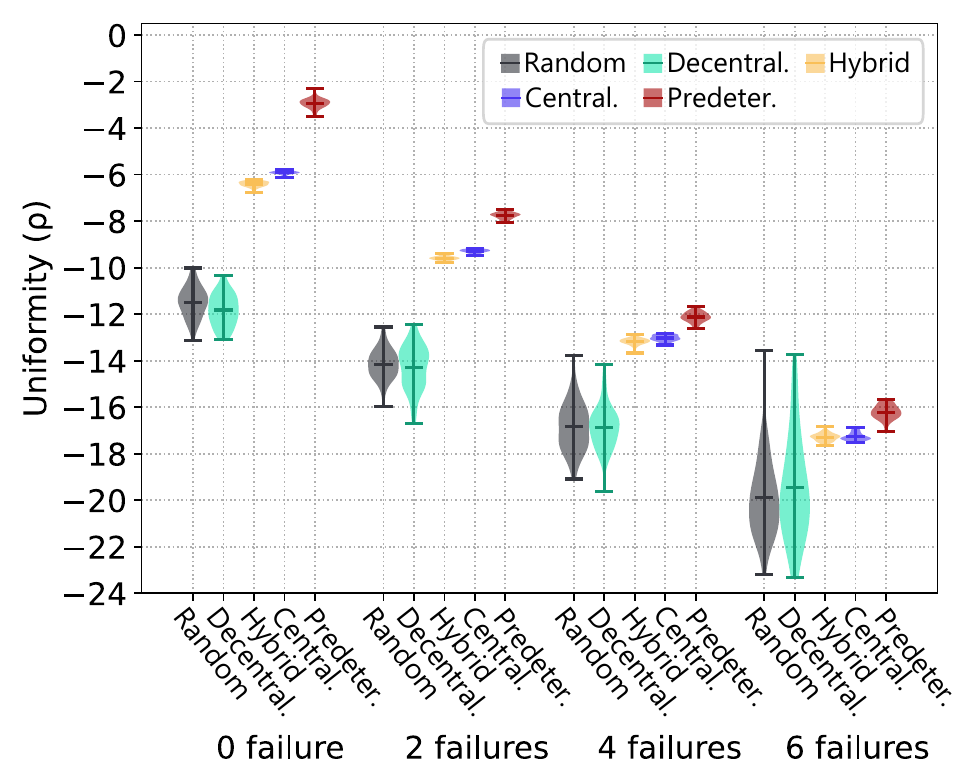}}
  \caption{\textbf{Fault tolerance:} Overall coverage performance under varying loss conditions. Random walk (gray), decentralized with beacons (green), hybrid formation (yellow), centralized formation (purple), and predetermined (red) approaches.}
  \label{fig:fault-overall}
\end{figure*}

\begin{figure*}[h]
  \centering
  \begin{minipage}[t]{0.42\textwidth}
        \subfigure[\label{fig:Scalability_Completeness}
         Scalability: Coverage completeness.]
          {\includegraphics[width=1.0\textwidth]{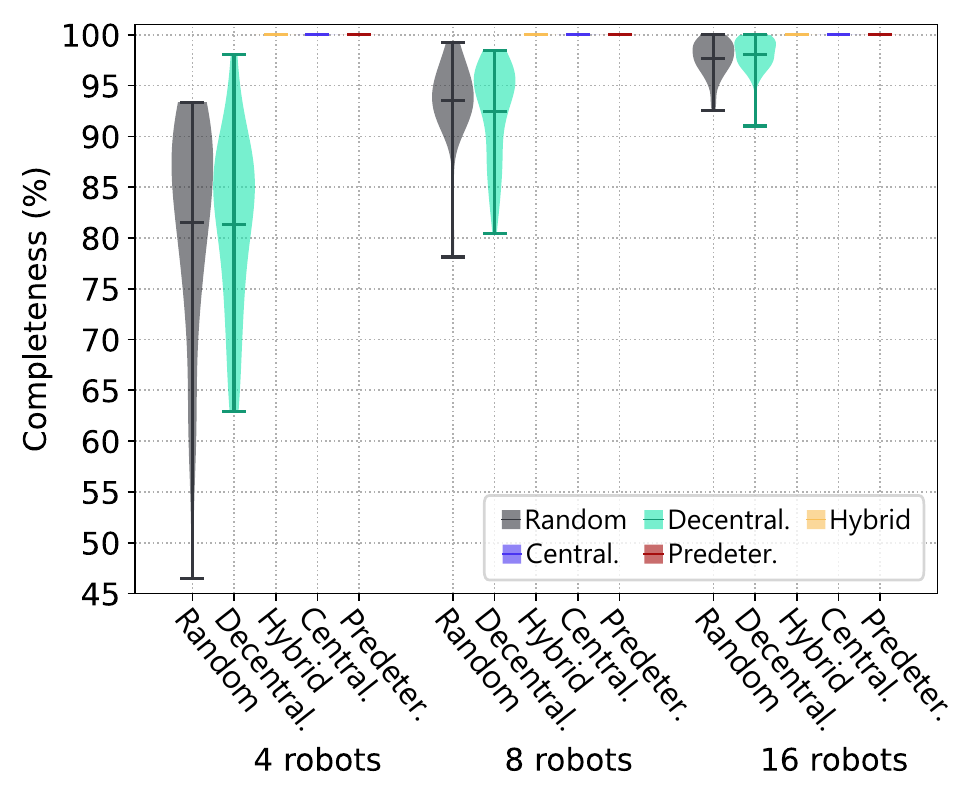}}
  \end{minipage}
  \begin{minipage}[t]{0.42\textwidth}
        \subfigure[\label{fig:Scalability_Uniformity}
         Scalability: Coverage uniformity.]
          {\includegraphics[width=1.0\textwidth]{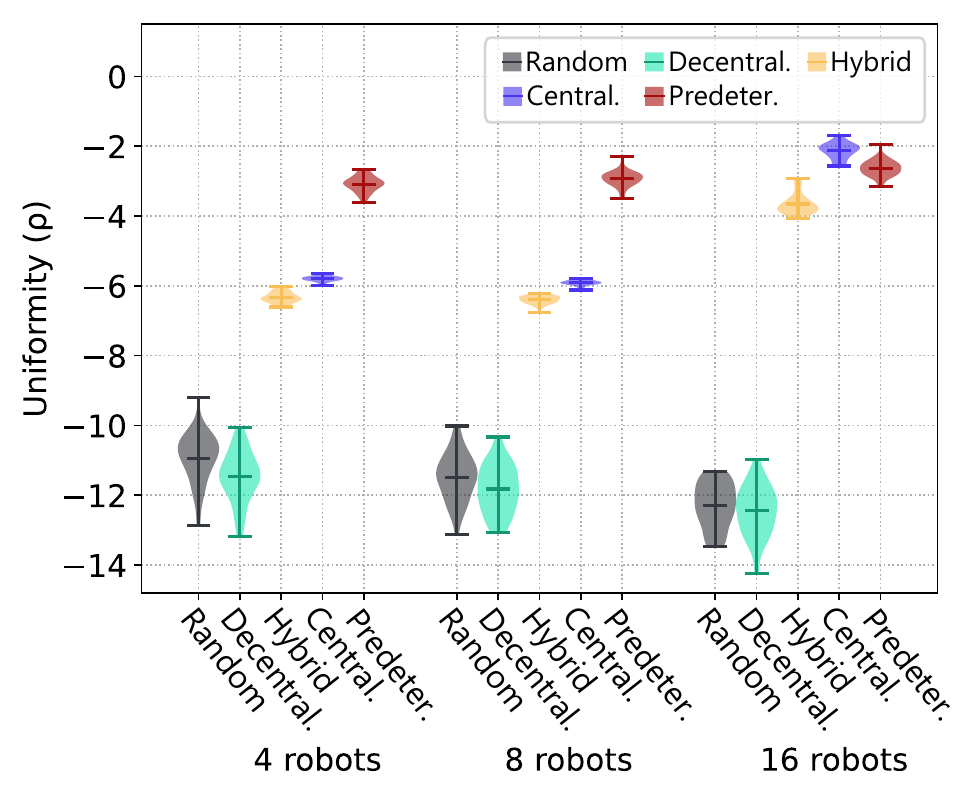}}
  \end{minipage}
    \begin{minipage}[t]{0.43\textwidth}
        \subfigure[\label{fig:Scalability_Time}
        Scalability: Sweep completion time.]
          {\includegraphics[width=1.0\textwidth]{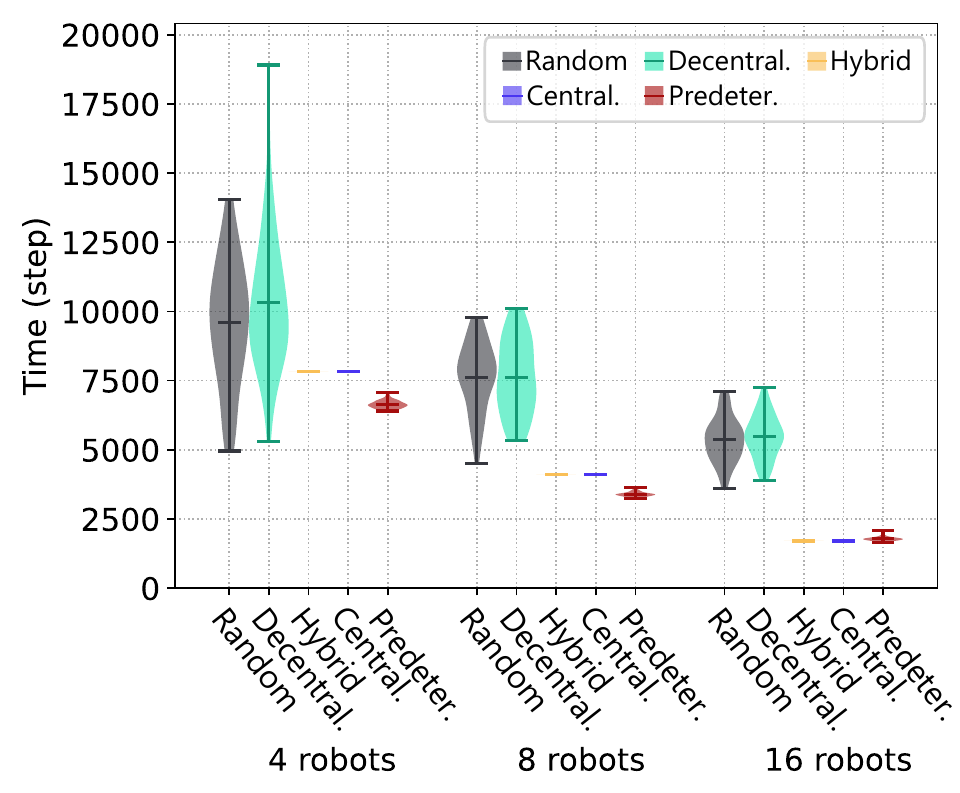}}
  \end{minipage}
  \caption{\textbf{Scalability:} Overall coverage performance and sweep completion time in systems with 4, 8, or 16 ground robots. Random walk (gray), decentralized with beacons (green), hybrid formation (yellow), centralized formation (purple), and predetermined (red) approaches.}
\label{fig:scalability}
\end{figure*}

\subsubsection{Coverage uniformity}

We define coverage uniformity as a measure of the variability of cell visit time.  
The best performing coverage approach would be one in which 1) all cells are visited and 2) they are all visited for an equal amount of time. In this ideal case, there would be no variability in cell visit time, resulting in a uniformity value of $0$.

For each trial, ${v}_i \in \mathbf{v}$ is defined as the total time spent by all robots in cell $i$. The coverage uniformity $\rho$ is the norm of $\mathbf{v}$, calculated as follows:
\begin{equation}
\label{eq:p}
\rho = -\sqrt{\sum_{i=1}^{c} |v_i - M(\mathbf{v})| \over c^{\small\textsc{visited}}},
\end{equation}
where $M(\mathbf{v})$ is the median of $\mathbf{v}$, $c$ is the number of cells, and $c^{\small\textsc{visited}}$ is the number of visited cells. The lower the value of $\rho$, the less uniformity between cells; the ideal, most uniform case is $\rho = 0$.

The coverage uniformity results are presented in Fig.~\ref{fig:perf-overall} as well as Tables 2 and 3 in the Appendix.

At the end of one sweep, the predetermined approach has noticeably more uniform coverage (higher $p$) than the centralized and hybrid formation approaches, which in turn have noticeably more uniform coverage than the two decentralized approaches, for all obstacle quantities (see Fig.~\ref{fig:perf-overall}b). 
All approaches have low variability of coverage uniformity between trials, with the formation approaches having the least.
In all approaches, coverage uniformity worsens slightly as the number of obstacles increases, with the predetermined approach worsening at a faster rate than the other approaches.

\subsection{Fault tolerance}

We assess fault tolerance in terms of the impact that robot failures have on coverage completeness and uniformity, and sweep completion time. 
We run the fault tolerance experiments in the setup with 100 obstacles and, at time step 400, impose failure on either two, four, or six of the eight total ground robots. When a ground robot fails, it stops moving, but it continues to be included in the calculation of coverage uniformity. Thus, a failed ground robot stops improving coverage completeness and meanwhile consistently worsens coverage uniformity.

The fault tolerance results show that both coverage completeness and coverage uniformity are heavily affected in all five approaches. Both completeness and uniformity worsen at a much faster rate in the predetermined and formation approaches than in the decentralized ones (see Fig.~\ref{fig:fault-overall}), with the predetermined approach worsening the most. 
However, in terms of coverage uniformity, although the decentralized approaches worsen at a much slower rate, they still under-perform the predetermined and formation approaches in all cases.

In terms of coverage completeness, the decentralized approaches outperform the predetermined and formation approaches only when running for a longer period. When compared based on elapsed time, the predetermined and formation approaches perform better or equivalently. That is, at the end of one sweep, the predetermined and formation approaches achieve worse coverage completeness than the two decentralized approaches, in all failure conditions (2, 4, and 6 failures), with the greatest gap occurring at 4 failures (approx. 80\% completeness for the decentralized approaches versus 60\% completeness for the other three). However, because of their adaptive termination condition, the decentralized approaches also take up to three times as long to complete a sweep in the fault tolerance experiments. When compared based on elapsed time (see Fig.~\ref{fig:fault-time}), the coverage completeness of the predetermined and formation approaches progressively nears that of the decentralized approaches as more robots fail (with the hybrid formation approach sometimes dipping below the decentralized approaches), until the point of 6 failures, when all five approaches perform equivalently.

\subsection{Scalability}

We assess scalability in terms of the impact of the number of ground robots on coverage completeness and uniformity as well as on sweep completion time (we ignore the number of UAVs in this assessment). 
We run the scalability experiments in the setup with 100 obstacles, with either four, eight, or 16 ground robots.

As the number of robots increases, the coverage completeness achieved by the predetermined and formation approaches stays the same (perfect or nearly perfect). At the same time, the completeness achieved by the decentralized approaches improves rapidly, nearing the perfect completeness of the other approaches in the system size of 16 robots (97.7\% to 98\% completeness). The sweep completion time of all approaches also improves as system size grows, with the predetermined and formation approaches improving at a slightly faster rate. The predetermined and formation approaches also improve in terms of coverage uniformity as the system size increases. The coverage uniformity of the decentralized approaches, by contrast, steadily worsens. This suggests that the better completeness and faster speeds of the decentralized approaches are achieved through increased redundancy.

\section{Discussion}

It is generally expected that fully centralized and predetermined approaches will have higher speed, efficiency, and accuracy, while fully decentralized approaches will have greater scalability, flexibility, and tolerance to robot failures.

As expected, our results show that the predetermined approach outperforms all other approaches in terms of both coverage completeness and coverage uniformity, and that the formation approaches outperform the decentralized ones. However, the decentralized approaches did not outperform the others as much as expected in terms of fault tolerance and scalability.

Regarding coverage completeness, the two decentralized approaches did not achieve coverage completeness as high as some existing results in the literature. (Note that coverage uniformity is not sufficiently covered in the literature to enable a comparison.) Our random walk approach (using random billiards) achieved up to an average of 93.5\% completeness (in the lowest difficulty environment, see Table 3 in the Appendix), while 97\% completeness was reported by~\cite{pang2021effect} using random walks. However, our experiments enforced a termination condition based on a tradeoff between completeness and uniformity, while uniformity was not considered in~\cite{pang2021effect}. In our preliminary tests (see Table 1 in the Appendix), when the random billiards approach was allowed to run for approximately double the time (30\,000 time steps), it achieved 96.2\% completeness, which is very similar to the results reported by~\cite{pang2021effect}. Therefore, we conclude that our random billiards approach is still adequately representative of the random walk category.

Our random billiards approach achieved completeness roughly equivalent to that of our decentralized approach with beacons, and in our preliminary tests (see Table 1 in the Appendix) likewise achieved completeness roughly equivalent to that of our stigmergic approach, which is contrary to what one might expect based on the literature.
We posit that this is related to two factors. Firstly, simple random walk seems to often be underestimated compared to other decentralized approaches: it is slow and inefficient, but when run for a long time, can reach quite high coverage completeness. Secondly, our decentralized approach with beacons might not have incorporated enough centralization to start seeing performance benefits. In our decentralized approach with beacons, robots still walk randomly when not reacting to the beacons, and robots simply avoid beacons rather than using them as a navigation guide to reach target areas. By contrast, in the beacon-based approach by~\cite{stirling2010energy} which achieved a mean completeness of 99.7\%, robots never revert to a random walk, instead deploying themselves as a temporary beacon network that guides robots to desired locations and allows the robots to progress through the environment together as a group. Therefore, in terms of (de)centralization, the beacon-based approach in~\cite{stirling2010energy} might be considered a midway point between our decentralized approach with beacons and our hybrid formation approach. 

Regarding fault tolerance, while the decentralized approaches achieved better coverage completeness under robot failures at the end of a sweep, this advantage disappears if completeness is considered in terms of elapsed time instead of sweep end. Also, the decentralized approaches never achieved better uniformity at the end of a sweep. Overall, the predetermined and formation approaches worsened at a faster rate, but their worsening mostly brought them down to a performance level close to that of the decentralized approaches. Given that the predetermined and formation approaches worsened at a faster rate, it is possible that they would under-perform the decentralized approaches in even worse conditions (i.e., worse than the 75\% failure rate tested here). However, it is important to note that if the predetermined and formation approaches were permitted to adapt their sweep trajectories during operation according to the number of remaining robots after failure, their fault tolerance would substantially improve. In short, we conclude that fault tolerance through redundancy, although present, did not give the decentralized approaches a meaningful advantage over the other approaches in the tested conditions. Of course, it is also important to note that the failure of UAVs was not tested, and that the predetermined, formation, and decentralized with beacons approaches would all be susceptible to this failure type, while the random walk approach would not.

Regarding scalability, the decentralized approaches achieved worse scalability overall than the other approaches. Firstly, as the number of robots increased, the decentralized approaches improved in terms of completeness while the other approaches did not, but this cannot be considered a scalability disadvantage for the predetermined and formation approaches because they had already achieved near-perfect completeness with the smallest number of robots and therefore could not improve further (also, they outperformed the decentralized approaches in all cases). Secondly, as the number of robots increased, all approaches improved at a similar rate in terms of sweep completion time, but the predetermined and formation approaches remained faster than the decentralized approaches in all cases. Finally, as the number of robots increased, uniformity improved for the predetermined and formation approaches but worsened for the decentralized approaches.

It is important to note that in the tested conditions, density increased as the number of robots increased, and that the decentralized approaches might have shown much better scalability if the area size increased as the number of robots increased, thus keeping density the same. For decentralized approaches, increasing the density of robots in an operating area often has a negative impact on efficiency, due both to increased interference between robots and to increased redundancy. This has been observed in existing decentralized approaches, in which adding more robots improves coverage completeness but also worsens repeated coverage~\citep{koenig2001terrain}, and is confirmed by our scalability results (i.e., completeness improved but uniformity worsened). Also, in existing hybrid approaches, it has not yet been demonstrated that adding elements of centralization will overcome the trade-off between completeness and repeated coverage~\citep[e.g., in the beacon-based approach in][, high coverage completeness is achieved but coverage uniformity is not considered]{stirling2010energy}. In our scalability results, the decentralized approach with beacons indeed suffered from this trade-off, but the hybrid formation approach overcame it, achieving both near-perfect completeness and drastically improved uniformity as system size increased.

Of course, it is also important to note that other scalability issues such as communication bottlenecks, computation bottlenecks, and latency were not tested, and that the predetermined and formation approaches would potentially be susceptible to these scalability issues (especially in much larger system sizes, e.g., of 500 robots), while the decentralized approaches would not. However, the SoNS used in the hybrid approach has been shown to be resistant to communication bottlenecks, computation bottlenecks, and latency issues in other mission types, in systems consisting of more than 100 robots~\citep{zhu2024self}. Finally, it should also be noted that the predetermined approach, as designed here, might be susceptible to further communication and sensing problems in much larger area sizes, as it might become difficult for one UAV to supervise the entire area at once.

Regarding flexibility in cluttered environments, existing literature suggests that although decentralized approaches are less efficient in open environments, they may be better suited to environments cluttered with unknown objects~\citep[cf.][]{almadhoun2019survey}. However, in the tested conditions, our results suggest that the predetermined and formation approaches were just as resilient to the unknown obstacle positions as the decentralized approaches, in terms of coverage completeness and uniformity. To make a future comparative assessment between centralization and decentralization in terms of flexibility, it would be important to test many other types of cluttered and complex environments.

\subsection{Future work}

In short, there are many factors related to scalability and fault tolerance that are not considered here (communication bottlenecks, failure of a supervisor robot, re-organization or re-planning in a centralized system that experiences failures), but it is notable that at least some aspects of performance might be more scalable and fault-tolerant in more centralized systems.
Overall, although we see clear performance advantages in the more centralized approaches, including in the scalability and fault tolerance setups, these advantages could be hard to maintain in centralized systems in practice, under conditions such as communication disturbances or failure of a supervisor robot. The SoNS approach, however, has been shown in other mission types to be resilient to such failures and disturbances~\citep{zhu2024self}. Future research is needed to define the best way to maintain the performance advantages observed in this study while also integrating resilience to other types of faults and scalability to much larger system sizes, as seen in~\citep{zhu2024self}, and to understand which of these findings are generalizable to other task types beyond sweep coverage.

\section{Conclusion}

In terms of coverage completeness, coverage uniformity, and completion time, the more centralized approaches outperformed the more decentralized approaches and the predetermined approach outperformed all other approaches\,---\,as expected. 
The performance of our random billiards approach showed that it is adequately representative of the random walk category in the literature, for the sweep coverage task. Our decentralized approach with beacons performed roughly equivalently to our random billiards approach, suggesting that greater centralization would need to be integrated to start seeing performance benefits.

In terms of fault tolerance, scalability, and flexibility, although the decentralized approaches outperformed the more centralized approaches by some metrics, they did not display the strong outperformance that had been expected. For example, the decentralized approaches achieved better coverage completeness under robot failures at the end of a sweep, but not at the end of the same elapsed time. In terms of coverage uniformity, although the performance of the more centralized approaches worsened at a much faster rate as more robots failed, they still always outperformed the decentralized approaches in the tested conditions. As another example, in terms of scalability, the coverage completeness of the decentralized approaches improved drastically as the number of robots increased, but still never outperformed the more centralized approaches, because the more centralized approaches always had near-perfect coverage completeness regardless of the number of robots. Also, the coverage uniformity of the more centralized approaches improved as the number of robots increased, while that of the decentralized approaches worsened. Finally, regarding flexibility in response to more cluttered environments, there was no notable difference between the decentralized and the more centralized approaches, in the tested conditions.

Overall, the more centralized approaches greatly outperformed the decentralized ones in terms of sweep coverage performance (coverage completeness, coverage uniformity, completion time). The decentralized approaches showed much better fault tolerance as well as better scalability by some metrics (i.e., slower overall performance degradation under increasing robot failures, and faster coverage completeness improvement under an increasing number of robots), but these fault tolerance and scalability benefits did not cause the decentralized approaches to outperform the more centralized ones\,---\,rather, the performance of the decentralized approaches became more similar to that of the centralized ones, in the tested conditions. However, the conclusions of this study have been demonstrated only for a specific set of experimental conditions. There are many additional conditions that should be studied in future work, such as communication bottlenecks, failure of a supervisor robot, more complex environments, and scalability to much larger system sizes.

\section*{Acknowledgements}
\textbf{Funding:} This work was partially supported by the Program of Concerted Research Actions (ARC) of the Universit{\'e} libre de Bruxelles; by the Ontario Trillium Scholarship Program through the University of Ottawa and the Government of Ontario, Canada; by the Office of Naval Research Global (Award N62909-19-1-2024); by the European Union's Horizon 2020 research and innovation programme under the Marie Sk\l{}odowska-Curie grant agreement No 846009; and by the China Scholarship Council (grant number 201706270186).
Mary Katherine Heinrich and Marco Dorigo acknowledge support from the Belgian F.R.S.-FNRS, of which they are a Postdoctoral Researcher and a Research Director respectively.

\noindent\textbf{Authors' contributions:}
AJ, MW, MA, MD, and MKH conceived and planned the experiments. AJ, MA, and WZ developed the code used in the experiments, supervised by MW, MA, and MKH. AJ conducted the experiments and collected the data. AJ, MW, MD, and MKH led the interpretation of the experimental results and conducted the analysis. The manuscript was written by AJ, MW, MD, and MKH, led by MKH. All authors read and approved the final manuscript.

\subsubsection*{Conflicts of interest}

The authors declare that they have no conflict of interest.

\subsubsection*{Availability of data and material}

All experimental data collected during the study are open-access, available on Zenodo: \url{https://doi.org/10.5281/zenodo.14846074}.

\subsubsection*{Code availability}

The code used in the experiments is open-source, available on GitHub: \url{https://github.com/BlueDiamond07/MultiRobotSweepCoverage}. It is also available on Zenodo: \url{https://doi.org/10.5281/zenodo.14846074}.

\bibliographystyle{spbasic}
\bibliography{mybibliography}

\input{appendices}

\end{document}

%% file: appendices.tex
\begin{table*}[h!]
\section*{Appendix: Experimental design}
\centering
\caption{Preliminary testing: comparison of coverage completeness between fully decentralized approaches (five different random walks and one stigmergic approach.}
\label{SO_1}
\begin{tabular}{|c|c|c|c|c|}
 \hline
  & {15\,770 steps} & {20\,000 steps} & {25\,000 steps} & {30\,000 steps} \\
 \hline
 {Correlated random walk} & 81,690 \%  & 88,326 \% & 93,018 \% & 95,478 \% \\
\hline
 {L\'{e}vy walk} & 80,628 \% & 87,072 \% & 92,236 \% & 94,882 \% \\
\hline
 {Brownian motion} & 73,268 \%  & 80,314 \% & 87,424 \% & 92,288 \% \\
\hline
 {Cauchy motion} & 48,176 \% & 56,57 \% & 57,94 \% & 72,036 \% \\
\hline
 {Random billiards} & 82,362 \% & 88,812 \% & 93,598 \%  & 96,204 \% \\
\hline
 {Stigmergic} & 82,258 \%  & 89,296 \% & 93,848 \% & 96,412 \% \\
\hline
\end{tabular}
\vspace{5mm}
\end{table*}

\begin{algorithm*}[h!]
\caption{{\bf -- Ground robot:} Predetermined (more detailed version).}
\label{alg:detailed-predetermined}
\begin{algorithmic}[1]
  \vspace{1mm}
  \If{an object is detected {\bf either} $\leq$ 3\,cm away and directly in front of the heading {\bf or} $\leq$ 1\,cm away and within 30$^{\circ}$ of the heading on the lefthand side}
  \vspace{1mm}
  \State \parbox[t]{420pt}{follow a motion routine resulting in a roughly half-circle trajectory around the object(s) {\bf until} the UAV sends a message indicating that the main path (i.e., the predetermined target sweep path) should be resumed\strut}
  \vspace{1mm}
  \ElsIf{an object is detected {\bf either} $\leq$ 3\,cm away and within 30$^{\circ}$ to 60$^{\circ}$ of the heading on the lefthand side {\bf or} $\leq$ 4\,cm away and within 60$^{\circ}$ of the heading on the righthand side}
  \vspace{1mm}
  \State \parbox[t]{420pt}{turn slightly away from the object while moving forward\strut}
  \vspace{1mm}
  \Else~follow motion instructions received from the UAV 
  \vspace{1mm}
  \EndIf
\end{algorithmic}
\end{algorithm*}

\clearpage
\begin{table*}[h!]
\section*{Appendix: Supplemental results}
\vspace{-10mm}
\small
\centering
\begin{subtable}\centering
\caption{Sweep coverage performance in all three obstacle setups: coverage completeness ($\alpha$), in \%; coverage uniformity ($\rho$); and sweep completion time ($T$), in time steps. The table reports the mean ({M}) and standard deviation ({SD}) for all three metrics, as well as the 95\% confidence interval ({CI}) for coverage completeness and uniformity.}
\label{tab:performance:obs}
\setlength\tabcolsep{2.4pt}
\begin{tabular}{|c|c|c|c|c|c|c|c|c|}
\hline
& {M($\alpha$)} & {SD($\alpha$)} & {CI($\alpha$)} & {M($\rho$)} & {SD($\rho$)} & {CI($\rho$)} & {M($T$)} & {SD($T$)}\\
\hline
{Random walk} & 92.30 & 4.40 & (91.59, 93.00)& -11.87 & 0.73 & (-11.99, -11.75) & 7701.78 & 1267.18\\ 
\hline
{Decentralized with beacons} & 91.48 & 5.30 & (90.64, 92.33) & -12.26 & 0.80 & (-12.39 -12.13) & 7817.43 & 1397.67\\ 
\hline
{Hybrid formation} & 99.91 & 0.52 & (99.83, 99.99) & -6.55 & 0.18 & (-6.58, -6.52) & 4101.00 & 0\\ 
\hline
{Centralized formation} & 100 & 0 & (100, 100) & -6.13 & 0.19 & (-6.16, -6.09) & 4101.00 & 0\\ 
\hline
{Predetermined} & 99.99 & 0.10 & (99.97, 100) & -3.54 & 0.55 & (-3.63, -3.45) & 3633.09 & 232.49\\ 
\hline
\end{tabular}
\end{subtable}
\end{table*}

\begin{table*}[h!]
\small
\centering
\begin{subtable}\centering
\caption{Sweep coverage performance in each separate obstacle setup (100, 200, and 300 obstacles): coverage completeness ($\alpha$), in \%; coverage uniformity ($\rho$); and sweep completion time ($T$), in time steps. The table reports the mean ({M}) and standard deviation ({SD}) for all three metrics, as well as the 95\% confidence interval ({CI}) for coverage completeness and uniformity.}
\label{tab:performance:obs100200300}
\setlength\tabcolsep{2.4pt}
\begin{tabular}{|c|c|c|c|c|c|c|c|c|}
\hline
 &  \multicolumn{8}{c|}{100 obstacles}\\
 \hline
 & {M($\alpha$)} & {SD($\alpha$)} & {CI($\alpha$)} & {M($\rho$)} & {SD($\rho$)} & {CI($\rho$)} & {M($T$)} & {SD($T$)}\\
\hline
{Random walk} & 93.52 & 4.02 & (92.40, 94.63) & -11.49 & 0.68 & (-11.68, -11.30) & 7624.46 & 1254.89\\ 
\hline
{Decentralized with beacons} & 92.43 & 4.75 & (91.11, 93.75) & -11.82 & 0.69 & (-12.01, -11.63) & 7612.14 & 1357.60 \\ 
\hline
{Hybrid formation} & 100 & 0 & (100, 100) & -6.40 & 0.12 & (-6.43, -6.37) & 4101.00 & 0\\ 
\hline
{Centralized formation} & 100 & 0 & (100, 100) & -5.91 & 0.07 & (-5.93, -5.89) & 4101.00 & 0\\ 
\hline
{Predetermined} & 100 & 0 & (100, 100) & -2.93 & 0.22 & (-2.99, -2.87) & 3391.60 & 81.84\\ 
\hline
\multicolumn{9}{c}{}\\
\hline
 &  \multicolumn{8}{c|}{200 obstacles}\\
\hline
& {M($\alpha$)} & {SD($\alpha$)} & {CI($\alpha$)} & {M($\rho$)} & {SD($\rho$)} & {CI($\rho$)} & {M($T$)} & {SD($T$)}\\
\hline
{Random walk} & 91.37 & 4.92 & (90.01, 92.74) & -11.94 & 0.60 & (-12.11, -11.78) & 7606.66 & 1430.19\\ 
\hline
{Decentralized with beacons} & 90.80 & 5.32 & (89.32, 92.27) & -12.34 & 0.76 & (-12.55, -12.13) & 7742.64 & 1359.02\\ 
\hline
{Hybrid formation} & 99.94 & 0.39 & (99.83, 100) & -6.56 & 0.12 & (-6.59, -6.52) & 4101.00 & 0\\ 
\hline
{Centralized formation} & 100 & 0 & (100, 100) & -6.13 & 0.09 & (-6.16, -6.11) & 4101.00 & 0\\ 
\hline
{Predetermined} & 100 & 0 & (100, 100) & -3.59 & 0.25 & (-3.66, -3.52) & 3653.10 & 136.82\\ 
\hline
\multicolumn{9}{c}{}\\
\hline
 &  \multicolumn{8}{c|}{300 obstacles}\\
\hline
& {M($\alpha$)} & {SD($\alpha$)} & {CI($\alpha$)} & {M($\rho$)} & {SD($\rho$)} & {CI($\rho$)} & {M($T$)} & {SD($T$)}\\
\hline
{Random walk} & 92.01 & 4.02 & (90.89, 93.12) & -12.18 & 0.75 & (-12.39, -11.97) & 7874.22 & 1103.14\\ 
\hline
{Decentralized with beacons} & 91.23 & 5.75 & (89.63, 92.82) & -12.63 & 0.74 & (-12.83, -12.42) & 8097.52 & 1456.51\\ 
\hline
{Hybrid formation} & 99.80 &  0.80 & (99.58, 100) & -6.70 & 0.15 & (-6.74, -6.66) & 4101.00 & 0\\ 
\hline
{Centralized formation} & 100 & 0 & (100, 100) & -6.33 & 0.09 & (-6.36, -6.31) & 4101.00 & 0\\ 
\hline
{Predetermined} & 99.97 & 0.17 & (99.92, 100) & -4.11 & 0.28 & (-4.19, -4.03) & 3854.58 & 170.19\\ 
\hline
\end{tabular}
\end{subtable}
\end{table*}

\begin{table*}[h!]
\small
\centering
\begin{subtable}\centering
\caption{Fault tolerance results: coverage completeness ($\alpha$), in \%; coverage uniformity ($\rho$); and sweep completion time ($T$), in time steps. The table reports the mean ({M}) and standard deviation ({SD}) for all three metrics, as well as the 95\% confidence interval ({CI}) for coverage completeness and uniformity.}
\label{tab:performance-change:fault}
\setlength\tabcolsep{2.4pt}
\begin{tabular}{|c|c|c|c|c|c|c|c|c|}
\hline
 &  \multicolumn{8}{c|}{\bf No failure}\\
\hline
& M($\alpha$) & SD($\alpha$) & CI($\alpha$) & M($\rho$) & SD($\rho$) & CI($\rho$) & M($T$) & SD($T$)\\
\hline
{Random walk} & 93.52 & 4.02 & (92.40, 94.63) & -11.49 & 0.68 & (-11.68, -11.30) & 7624.46 & 1254.89\\ 
\hline
{Decentralized with beacons} & 92.43 & 4.75 & (91.11, 93.75) & -11.82 & 0.69 & (-12.01, -11.63) & 7612.14 & 1357.60\\ 
\hline
{Hybrid formation} & 100 & 0 & (100, 100) & -6.40 & 0.12 & (-6.43, -6.37) & 4101.00 & 0\\ 
\hline
{Centralized formation} & 100 & 0 & (100, 100) & -5.91 & 0.07 & (-5.93, -5.89) & 4101.00 & 0\\ 
\hline
{Predetermined} & 100 & 0 & (100, 100) & -2.93 & 0.22 & (-2.99, -2.87) & 3391.60 & 81.84\\ 
\hline
\multicolumn{9}{c}{}\\
\hline
 &  \multicolumn{8}{c|}{\bf 2 failures}\\
\hline
& M($\alpha$) & SD($\alpha$) & CI($\alpha$) & M($\rho$) & SD($\rho$) & CI($\rho$) & M($T$) & SD($T$)\\
\hline
{Random walk} & 90.91 & 5.19 & (89.47, 92.35) & -14.15 & 0.66 & (-14.33, -13.97) & 8887.28 & 1605.69\\
\hline
{Decentralized with beacons} & 88.78 & 5.65 & (87.22, 90.35) & -14.31 & 0.96 & (-14.58, -14.04) & 8530.18 & 1663.75\\
\hline
{Hybrid formation} & 80.98 & 0.35 & (80.89, 81.08) & -9.59 & 0.08 & (-9.62, -9.57) & 4101.00 & 0\\
\hline
{Centralized formation} & 80.98 & 0.26 & (80.90, 81.05) & -9.28 &  0.075 & (-9.31, -9.26) & 4101.00 & 0\\
\hline
{Predetermined} & 79.38 & 0.31 & (79.29, 79.46) & -7.75 & 0.13 & (-7.79, -7.72) & 3387.58 & 84.89\\
\hline
\multicolumn{9}{c}{}\\
\hline
 &  \multicolumn{8}{c|}{\bf 4 failures}\\
\hline
& M($\alpha$) & SD($\alpha$) & CI($\alpha$) & M($\rho$) & SD($\rho$) & CI($\rho$) & M($T$) & SD($T$)\\
\hline
{Random walk} & 81.97 & 8.98 & (79.48, 84.46) & -16.84 &  1.05 & (-17.13, -16.55) & 9275.08 & 2351.96\\
\hline
{Decentralized with beacons} & 80.45 & 8.71 & (78.03, 82.86) & -16.87 & 0.96 & (-17.14, -16.60) & 8900.72 & 1947.66\\
\hline
{Hybrid formation} & 61.66 & 0.84 & (61.42, 61.89) & -13.18 & 0.15 & (-13.22, -13.14) & 4101.00 & 0\\
\hline
{Centralized formation} & 61.74 & 0.52 & (61.60, 61.89) & -13.02 & 0.13 & (-13.06, -12.99) & 4101.00 & 0\\
\hline
{Predetermined} & 58.59 & 0.50 & (58.45, 58.73) & -12.12 &  0.22 & (-12.18, -12.06) & 3360.46 & 125.67\\
\hline
\multicolumn{9}{c}{}\\
\hline
 &  \multicolumn{8}{c|}{\bf 6 failures}\\
\hline
& M($\alpha$) & SD($\alpha$) & CI($\alpha$) & M($\rho$) & SD($\rho$) & CI($\rho$) & M($T$) & SD($T$)\\
\hline
{Random walk} & 60.55 & 13.25 & (56.87, 64.22) & -19.90 & 1.69 & (-20.37, -19.43) & 8316.20 & 2921.84\\
\hline
{Decentralized with beacons} & 53.55 & 15.98 & (49.13, 57.98) & -19.47 & 2.24 & (-20.09, -18.84) & 7186.90 & 3376.36\\
\hline
{Hybrid formation} & 41.76 & 0.85 & (41.52, 41.99) & -17.29 & 0.17 & (-17.34, -17.24) & 4101.00 & 0\\
\hline
{Centralized formation} & 41.94 & 0.80 & (41.72, 42.16) & -17.27 & 0.16 & (-17.31, -17.22) & 4101.00 & 0\\
\hline
{Predetermined} & 38.05 & 0.56 & (37.89, 38.20) & -16.23 & 0.31 & (-16.31, -16.14) & 3261.78 & 124.83\\
\hline
\end{tabular}
\end{subtable}
\end{table*}
\begin{table*}[h!]
\vspace{-5mm}
\small
\centering
\begin{subtable}\centering
\caption{Fault tolerance results for different failure rates compared to no failure: change ($\Delta$) in mean (M) w.r.t. 0 failures for coverage completeness ($\alpha$), in \%; coverage uniformity ($\rho$); and sweep completion time ($T$), in time steps.}
\label{tab:fault_tolerance-performance-change}
\begin{tabular}{|c|c|c|c|}
\hline
 &  \multicolumn{3}{c|}{\bf Change in mean for 2 failures}\\
\hline
&$\Delta$ M($\alpha$) & $\Delta$ M($\rho$) & $\Delta$ M($T$)\\
\hline
{Random walk} & -2.61 & -2.66 & +1262.82\\
\hline
{Decentralized with beacons} & -3.65 & -2.49 & +918.04\\
\hline
{Hybrid formation} & -19.02 & -3.19 & 0\\
\hline
{Centralized formation} & -19.02 & -3.37 & 0\\
\hline
{Predetermined} & -20.62 & -4.82 & -4.02\\
\hline
\multicolumn{4}{c}{}\\
\hline
 & \multicolumn{3}{c|}{\bf Change in mean for 4 failures}\\
\hline
&$\Delta$ M($\alpha$) & $\Delta$ M($\rho$) & $\Delta$ M($T$)\\
\hline
{Random walk} & -11.55 & -5.35 & +1650.62\\
\hline
{Decentralized with beacons} & -11.98 & -5.05 & +1288.58\\
\hline
{Hybrid formation} & -38.34 & -6.78 & 0\\
\hline
{Centralized formation} & -38.26 & -7.11 & 0\\
\hline
{Predetermined} & -41.41 & -9.19 & -31.14\\
\hline
\multicolumn{4}{c}{}\\
\hline
 & \multicolumn{3}{c|}{\bf Change in mean for 6 failures}\\
\hline
&$\Delta$ M($\alpha$) & $\Delta$ M($\rho$) & $\Delta$ M($T$)\\
\hline
{Random walk} & -32.97 & -8.41 & +691.74\\
\hline
{Decentralized with beacons} & -38.88 & -7.65 & -425.24\\
\hline
{Hybrid formation} & -58.24 & -10.89 & 0\\
\hline
{Centralized formation} & -58.06 & -11.36 & 0\\
\hline
{Predetermined} & -61.95 & -13.30 & -129.82\\
\hline
\end{tabular}
\end{subtable}
\end{table*}

\begin{table*}
\small
\centering
\begin{subtable}\centering
\caption{Scalability results: coverage completeness ($\alpha$), in \%; coverage uniformity ($\rho$); and sweep completion time ($T$), in time steps. The table reports the mean ({M}) for all three metrics, as well as the standard deviation ({SD}), 95\% confidence interval ({CI}), and mean per robot for coverage completeness and uniformity.}
\label{tab:performance:scal}
\setlength\tabcolsep{2.4pt}
\begin{tabular}{|c|c|c|c|c|c|c|c|c|c|}
\hline
 &  \multicolumn{9}{c|}{\bf 4 robots}\\
\hline
 &&&&{M($\alpha$)}&&&&{M($\rho$)}&\\
& {M($\alpha$)} & {SD($\alpha$)} & {CI($\alpha$)} & {/robot} & {M($\rho$)} & {SD($\rho$)} & {CI($\rho$)} & {/robot} & {M($T$)}\\
\hline
Random walk & 81.55 & 10.80 & (78.56, 84.55) & 20.39 & -10.94 & 0.71 & (-11.14, -10.74) & -2.74 & 9602.12\\
\hline
Decentralized with beacons & 81.31 & 8.64 & (78.92, 83.71) & 20.33 & -11.48 & 0.73 & (-11.68, -11.27) & -2.87 & 10329.52\\
\hline
Hybrid formation & 100 & 0 & (100, 100) & 25 & -6.33 & 0.15 & (-6.37, -6.29) & -1.58 & 7840.00\\
\hline
Centralized formation & 100 & 0 & (100, 100) & 25 & -5.79 & 0.07 & (-5.81, -5.78) & -1.45 & 7840.00\\
\hline
Predetermined & 100 & 0 & (100, 100) & 25 & -3.09 & 0.22 & (-3.15, -3.03) & -0.77 & 6642.84\\
\hline
\multicolumn{9}{c}{}\\
\hline
 &  \multicolumn{9}{c|}{\bf 8 robots}\\
 \hline
  &&&&{M($\alpha$)}&&&&{M($\rho$)}&\\
& {M($\alpha$)} & {SD($\alpha$)} & {CI($\alpha$)} & {/robot} & {M($\rho$)} & {SD($\rho$)} & {CI($\rho$)} & {/robot} & {M($T$)}\\
 \hline
Random walk & 93.52 & 4.02 & (92.40, 94.63) & 11.69 & -11.49 & 0.68 & (-11.68, -11.30) & -1.44 & 7624.46\\
\hline
Decentralized with beacons & 92.43 & 4.75 & (91.11, 93.75) & 11.55 & -11.82 & 0.69 & (-12.01, -11.63) & -1.48 & 7612.14\\
\hline
Hybrid formation & 100 & 0 & (100, 100) & 12.50 & -6.40 & 0.12 & (-6.43, -6.37) & -0.80 & 4101.00\\
\hline
Centralized formation & 100 & 0 & (100, 100) & 12.50 & -5.91 & 0.07 & (-5.93, -5.89) & -0.74 & 4101.00\\
\hline
Predetermined & 100 & 0 & (100, 100) & 12.50 & -2.93 & 0.22 & (-2.99, -2.87) & -0.37 & 3391.60\\
\hline
\multicolumn{9}{c}{}\\
\hline
 &  \multicolumn{9}{c|}{\bf 16 robots}\\
 \hline
   &&&&{M($\alpha$)}&&&&{M($\rho$)}&\\
& {M($\alpha$)} & {SD($\alpha$)} & {CI($\alpha$)} & {/robot} & {M($\rho$)} & {SD($\rho$)} & {CI($\rho$)} & {/robot} & {M($T$)}\\
 \hline
Random walk & 97.67 & 1.75 & (97.19, 98.16) & 6.10 & -12.29 & 0.64 & (-12.47, -12.12) & -0.77 & 5355.06\\
\hline
Decentralized with beacons & 98.09 & 1.63 & (97.64, 98.54) & 6.13 & -12.45 & 0.71 & (-12.64, -12.25) & -0.78 & 5482.44\\
\hline
Hybrid formation & 100 & 0 & (100, 100) & 6.25 & -3.65 & 0.29 & (-3.74, -3.57) & -0.23 & 1702.00\\
\hline
Centralized formation & 100 & 0 & (100, 100) & 6.25 & -2.12 & 0.22 & (-2.18, -2.06) & -0.13 & 1702.00\\
\hline
Predetermined & 100 & 0 & (100, 100) & 6.25 & -2.63 & 0.23 & (-2.70, -2.57) & -0.16 & 1778.32\\
\hline
\end{tabular}
\end{subtable}
\end{table*}
\begin{table*}
\small
\centering
\begin{subtable}\centering
\caption{Scalability results: change ($\Delta$) in mean (M) for coverage completeness ($\alpha$) and coverage completeness per robot, in \%; coverage uniformity ($\rho$) and coverage uniformity per robot; and sweep completion time ($T$), in time steps.}
\label{tab:performance-change:scal}
\begin{tabular}{|c|c|c|c|c|c|}
\hline
 &  \multicolumn{5}{c|}{\bf Change in mean from 4 to 8 robots}\\
\hline
&$\Delta$ M($\alpha$) & $\Delta$ M($\alpha$) /robot & $\Delta$ M($\rho$) & $\Delta$ M($\rho$) /robot & $\Delta$ M($T$)\\
\hline
{Random walk} & +11.97 & -8.70 & -0.55 & +1.30 & -1977.66\\
\hline
{Decentralized with beacons} & +11.12 & -8.78 & -0.34 & +1.39 & -2717.38\\
\hline
{Hybrid formation} & 0 & -12.50 & -0.07 & +0.78 & -3739.00\\
\hline
{Centralized formation} & 0 & -12.50 & -0.12 & +0.71 & -3739.00\\
\hline
{Predetermined} & 0 & -12.50 & +0.16 & +0.40 & -3251.24\\
\hline
\multicolumn{6}{c}{}\\
\hline
 & \multicolumn{5}{c|}{\bf Change in mean from 8 to 16 robots}\\
\hline
&$\Delta$ M($\alpha$) & $\Delta$ M($\alpha$) /robot & $\Delta$ M($\rho$) & $\Delta$ M($\rho$) /robot & $\Delta$ M($T$)\\
\hline
{Random walk} & +4.15 & -5.59 & -0.80 & +0.67 & -2269.40\\
\hline
{Decentralized with beacons} & +5.66 & -5.42 & -0.63 & +0.70 & -2129.70\\
\hline
{Hybrid formation} & 0 & -6.25 & +2.75 & +0.57 & -2399.00\\
\hline
{Centralized formation} & 0 & -6.25 & +3.79 & +0.61 & -2399.00\\
\hline
{Predetermined} & 0 & -6.25 & +0.30 & +0.21 & -1613.28\\
\hline
\end{tabular}
\end{subtable}
\end{table*}